\documentclass[dvipsnames,format=sigconf,anonymous=false,review=false]{acmart}
\AtBeginDocument{%
  }

\copyrightyear{2025}
\acmYear{2025}
\setcopyright{rightsretained}
\acmConference[{\upshape This is the author's version of the work. It is posted here for your personal use. Not for redistribution. The definitive Version of Record was published in} GECCO '25]{Genetic and Evolutionary Computation Conference}{July 14--18, 2025}{Malaga, Spain}
\acmBooktitle{Genetic and Evolutionary Computation Conference (GECCO '25), July 14--18, 2025, Malaga, Spain}
\acmDOI{10.1145/3712256.3726345}
\acmISBN{979-8-4007-1465-8/2025/07}

\usepackage{graphicx}
\usepackage{algorithm}
\usepackage{algorithmic}
\usepackage{multirow}
\usepackage{array}
\usepackage{subcaption}
\usepackage{graphicx}
\usepackage{booktabs}
\usepackage{amsmath}
\usepackage{mathtools}
\usepackage{stmaryrd}
\usepackage{bbm}
\usepackage{dsfont}

\providecommand{\R}{\mathbb{R}} 
\providecommand{\E}{\mathbb{E}}

\providecommand{\ind}[1]{\ensuremath{\mathds{1}\left\{#1\right\}}}
\renewcommand{\geq}{\geqslant}
\renewcommand{\leq}{\leqslant}
\DeclarePairedDelimiterX{\inner}[2]{\langle}{\rangle}{#1, #2}

\DeclarePairedDelimiter{\abs}{\lvert}{\rvert}

\usepackage{cleveref}

\begin{document}

\title{Feature selection based on cluster assumption in PU learning}

\author{Motonobu Uchikoshi}
\orcid{0009-0009-6234-4734}
\affiliation{%
  \institution{The Japan Research Institute, Limited}
  \city{Shinagawa-ku}
  \state{Tokyo}
  \country{Japan}}
\email{uchikoshi.motonobu@jri.co.jp}

\author{Youhei Akimoto}
\orcid{0000-0003-2760-8123}
\affiliation{%
  \institution{University of Tsukuba \& RIKEN AIP}
  \city{Tsukuba}
  \state{Ibaraki}
  \country{Japan}}
\email{akimoto@cs.tsukuba.ac.jp}


\begin{abstract}
Feature selection is essential for efficient data mining and sometimes encounters the positive-unlabeled (PU) learning scenario, where only a few positive labels are available, while most data remains unlabeled. 
In certain real-world PU learning tasks, data subjected to adequate feature selection often form clusters with concentrated positive labels.
Conventional feature selection methods that treat unlabeled data as negative may fail to capture the statistical characteristics of positive data in such scenarios, leading to suboptimal performance. 
To address this, we propose a novel feature selection method based on the cluster assumption in PU learning, called FSCPU. 
FSCPU formulates the feature selection problem as a binary optimization task, with an objective function explicitly designed to incorporate the cluster assumption in the PU learning setting.
Experiments on synthetic datasets demonstrate the effectiveness of FSCPU across various data conditions. 
Moreover, comparisons with 10 conventional algorithms on three open datasets show that FSCPU achieves competitive performance in downstream classification tasks, even when the cluster assumption does not strictly hold.
\end{abstract}




\maketitle

\section{Introduction}
Positive-unlabeled (PU) learning~\cite{PUsurvey, semi_JMI_b21} is a special case of semi-supervised learning, wherein only a small amount of data are positively labeled, while most data are unlabeled. 
It frequently appears in real-world applications, such as anomaly detection, 
in which some malicious data are reported (i.e., labeled as positive) but other data are not labeled.

Feature selection is fundamental in PU learning because it enhances interpretability, model performance on downstream tasks, and computational efficiency.
However, feature selection in PU learning setting has been scarcely investigated in the literature, as compared to supervised and unsupervised learning settings.
A simple approach to feature selection tasks in PU learning is to treat all unlabeled data as negative and adopt a supervised feature selection approach~\cite{semi_JMI_b21}, or to ignore all positive labels and adopt an unsupervised feature selection approach. 
Although these approaches are helpful, their performance on downstream tasks is often insufficient, possibly owing to the biased labeling and the ignorance of the label information.

We aimed to develop a novel feature selection method for the PU-learning setting so as to incorporate the partial label information effectively. 
Effective utilization of the partial label information necessitates an adequate assumption on data and the development of a criterion for feature selection that reflects the posed assumption. 

Accordingly, we devised a novel cluster assumption for PU learning and formulated feature selection criteria based on the assumption.
The conventional cluster assumption \cite{sSelect_b4} states that, in a binary semi-supervised context, 
clustering data with appropriate feature selection would form the positive cluster and the negative one.
This study is based on the cluster assumption that positive clusters would form if data are appropriately clustered after applying suitable feature selection and selecting the correct clusters.
This suggests that positive data may originate from multiple classes and that the entire dataset can be divided into positive and remaining data.
The proposed feature selection criterion is built on this new cluster assumption and evaluates each feature selection pattern by using a cluster-by-cluster predictive performance metrics framework.
We employed an evolutionary approach to optimize this criterion.
To prevent the computational cost for the proposed criterion from being expensive during optimization, we introduced a constraint handling mechanism that forced each candidate solution generated during optimization to satisfy the cost bound. 
Finally, we proposed a novel Feature Selection approach based on Cluster assumption in PU learning, 
named \emph{FSCPU}.

We conducted experiments using artificial data to test whether the indicators and search methods based 
on the afore-described natural assumptions differ from existing indicators. 
The results showed that FSCPU performed better than existing approaches when the assumptions held. 
Moreover, to confirm its effectiveness in real-world task-like situations, we compared the performance of FSCPU with various baseline approaches---including supervised, unsupervised, semi-supervised, and reinforcement learning approaches---and tested it on certain open datasets by evaluating the classification performance of downstream models.

Our contributions are summarized as follows:
(I) We propose FSCPU, a novel optimization framework for feature selection in PU learning. 
FSCPU incorporates a newly introduced cluster assumption, formulating an objective function for PU learning.
(II) Experiments on artificial datasets, comparing FSCPU with $9$ baseline methods across $10$ different data conditions, reveal that FSCPU demonstrates advantages over conventional methods when at least the following conditions are met: (1) the cluster assumption holds, (2) the positive labeling rate is low, and (3) no interfering data hinder the clustering of positive instances.
(III) We assess the real-world applicability of FSCPU through experiments on three open datasets with ten baseline methods, showing that FSCPU may achieve competitive performance even on datasets where the validity of the cluster assumption is uncertain. 

\section{Related work}
Feature selection methods can be broadly categorized into filter, wrapper, and embedded methods \cite{semi_survey_b2}. 
Embedded methods incorporate feature selection within the learning process. 
There are several representative methods, including supervised approaches such as Lasso \cite{LASSO_b6} 
and semi-supervised methods that use support vector machines \cite{Xu2009DiscriminativeSF} or graph Laplacians \cite{JSFS_b3}.
Wrapper methods entail the use of learning models to evaluate feature subsets. 
Lightweight models such as DT-RFE \cite{DT-RFE_b7} are commonly employed, along with evolutionary algorithms 
and reinforcement learning methods. 
In this regard, evolutionary algorithms such as genetic algorithms (GAs) \cite{Gheyas2010FeatureSS} and particle swarm optimization (PSO) \cite{Xue2013ParticleSO} are widely used.
Reinforcement learning approaches can be categorized into multi-agent \cite{Multiagent-agent_RL_1_b14, Multiagent-agent_RL_2_b15} systems, wherein an agent is assigned to each feature, and 
single-agent systems \cite{ES-MCRFS_single_agent_b9, single-agent_RL_1_b13, single-agent_RL_2_b16}, wherein all features are handled by a single agent. 
Filter methods are independent of specific learning models and evaluate features based on intrinsic data properties. 
The proposed method falls within the filter method.

Filter methods include supervised, semi-supervised, and unsupervised techniques. 
Supervised filter methods such as K-Best \cite{K-best_b5} and mRMR \cite{mRMR_b8} are representative methods. 
When applied to PU learning through preprocessing (e.g., treating positive labels as 1 and unlabeled data as 0), supervised filter methods may suffer from performance degradation owing to the limited number of positive labels \cite{semi_survey_b2}.
Unsupervised filter methods, such as MRMR-UFS \cite{MRMR-UFS_b10}, cannot leverage label information in the PU learning setting, as they operate without considering the label data.
Semi-supervised filter-based feature selection methods typically assign scores based on structural assumptions regarding the data.
For instance, some methods rely on structural assumptions such as cluster assumption \cite{sSelect_b4} and manifold assumptions \cite{manifold_assumptions_b17}.
These approaches address the small number of labels available but do not assume a PU learning setting with only one type of label.
The number of feature selection methods specifically designed for PU learning is limited. 
An example of such a method is semi-JMI \cite{semi_JMI_b21}, which explicitly considers the PU learning setting. 
However, semi-JMI can suffer from performance degradation when a significant portion of positive labels is hidden within the unlabeled data.
The proposed method adopts a novel clustering assumption tailored for PU learning, which helps mitigate the issue of hidden positive labels among negative labels. 
By evaluating feature relevance at the cluster level rather than the individual instance level, the proposed method is expected to exhibit greater robustness in the PU learning setting. 

\providecommand{\R}{\mathds{R}}
\providecommand{\xx}{\boldsymbol{x}}
\section{Problem Setting}

Our problem is formulated as a feature selection under a positive-unlabeled (PU) learning setting. 
Let $\xx \in \mathcal{X} \subseteq \R^D$ represent the input data and $\phi: \mathcal{X} \to \R^d$ be the feature selector, where $d < D$ is the number of selected features. 
Our dataset consists of a set of positively-labeled data, $\mathcal{D}_{L} = \{(\xx_{i}, y_i=1)\}_{i=1}^{L}$, and a set of unlabeled data, $\mathcal{D}_{U} = \{\xx_{i}\}_{i=L+1}^{L+U}$.
That is, we have a relatively small amount of positively labeled data and a possibly large amount of unlabeled data as the training data. 

Our goal is to obtain a feature selection $\phi$ that maximizes the performance of downstream tasks.
We consider classification tasks as the downstream tasks. In particular, we consider two-class classification tasks such as anomaly detection tasks.
Let $f_\phi$ be a classifier for the downstream task trained on the datasets $\mathcal{D}_L^\phi = \{(\phi(\xx_i), y_i) : (\xx_i, y_i) \in \mathcal{D}_L\}$ and $\mathcal{D}_U^\phi = \{\phi(\xx_i) : \xx_i\in \mathcal{D}_U\}$ after the application of the feature selection $\phi$. 
Then, our objective is to locate the feature selection $\phi$ that maximizes the performance of a classifier trained using $\mathcal{D}_L^\phi$ and $\mathcal{D}_U^\phi$ for the downstream task.

\section{Proposed Method}

We propose a novel Feature Selection method based on Cluster assumption in PU learning, named \emph{FSCPU}.
FSCPU employs a filter approach in feature selection methods, segregates upstream feature selection from downstream classification methods, and focuses solely on feature selection.
In Section~\ref{sec:obj}, we propose a novel objective function based on our cluster assumption.
In Section~\ref{sec:cga}, we introduce a compact genetic algorithm as a search method and propose a constraint-handling technique specialized for our feature selection tasks.

\subsection{Objective Function for Feature Selection}\label{sec:obj}

First, we design an objective function $f(\phi)$ for feature selection pattern $\phi$ in the PU learning setting. 

Our objective function is designed based on the cluster assumption: \emph{if the data reflecting the optimal feature selection pattern are appropriately clustered, the discriminating targets are gathered in some clusters}.
It is mathematically described as follows.
First, to present our idea, we suppose that we have access to the set $\mathcal{D}_* = \{(\xx_i, y_i)\}$ of fully labeled data. 
Let $\mathcal{D}^\phi_* = \{\phi(\xx_i) : (\xx_i, y_i) \in \mathcal{D}_*\}$ be the set of input data in $\mathcal{D}_*$ after the feature selection by $\phi$.
Let $c: \mathcal{D}^\phi_* \mapsto (\mathcal{C}_1, \dots, \mathcal{C}_K)$ be a clustering method, where $\mathcal{C}_k$ represents the $k$th cluster. 
Then, the cluster assumption states that, under the optimal feature selection pattern, the clusters are split into two groups, $\mathcal{K}_* \subseteq \llbracket 1, K \rrbracket$ and $\mathcal{K}_*^c = \llbracket 1, K \rrbracket \setminus \mathcal{K}_*$, where
\begin{align}
\phi(\xx_i) \in \bigcup_{k \in \mathcal{K}_*} \mathcal{C}_{k} &\implies y_i = 1 ,\\
\phi(\xx_i) \in \bigcup_{k \in \mathcal{K}_*^c} \mathcal{C}_{k} &\implies y_i = 0.
\end{align}
That is, some clusters (indexed by $k \in \mathcal{K}_*$) are composed of the positively labeled data and the others are composed of the negatively labeled data.
In practice, the above conditions are too strict, but we expect that the probability of satisfying these conditions is sufficiently high. 

Based on this hypothesis, the objective function is ideally formulated as follows. 
Let $\mathcal{D}_{*,p}^\phi = \{\phi(\xx_i) : (\xx_i, y_i) \in \mathcal{D}^* \text{ and } y_i = 1\}$ be the set of positively labeled data in $\mathcal{D}_*$ after the feature selection by $\phi$.
Then, the objective function is defined as
\begin{align}
f_*(\phi) = \max_{\mathcal{K} \in 2^{\llbracket 1, K\rrbracket}} \underbrace{ \frac{ \abs{ \mathcal{D}_{*,p}^\phi \cap \bigcup_{k \in \mathcal{K} } \mathcal{C}_k}  }{ \abs{\mathcal{D}_{*,p}^\phi}} }_{\text{recall}} \underbrace{ \frac{ \abs{ \mathcal{D}_{*,p}^\phi \cap \bigcup_{k \in \mathcal{K} } \mathcal{C}_k}  }{ \abs{\mathcal{C}_k} } }_{\text{precision}}.\label{eq:f*}
\end{align}
To provide the interpretation of this objective, let $g_{\phi, \mathcal{K}}$ be the discriminator defined by the clustering method $c$, feature selection pattern $\phi$, and the subset $\mathcal{K}$ of cluster indices that are considered as positive.
That is,
\begin{equation}
g_{\phi, \mathcal{K}}(\xx_i) = \begin{cases}
1 & \xx_i \in \bigcup_{k \in \mathcal{K}} \mathcal{C}_{k},\\
0 & \xx_i \notin \bigcup_{k \in \mathcal{K}} \mathcal{C}_{k}.
\end{cases}
\end{equation}
Then, the first term of \eqref{eq:f*} is regarded as the recall of $g_{\phi, \mathcal{K}}$ and the second term is regarded as the precision of $g_{\phi, \mathcal{K}}$. The subset $\mathcal{K}$ is selected to maximize the product of the recall and the precision. 
Let $\mathcal{K}_*^{\phi}$ denote the optimal subset under $\phi$. 
Then, the optimal feature selection pattern in terms of $f_*$ is regarded as the one that results in the optimal discriminator $g_{\phi, \mathcal{K}_*^{\phi}}$ in terms of the product of the recall and the precision.

Unfortunately, we can not compute the objective function \eqref{eq:f*} because we have only the positively-labeled dataset $\mathcal{D}_L$ and the unlabeled dataset $\mathcal{D}_U$. To address this issue, we treat the unlabeled data in $\mathcal{D}_U$ as negatively-labeled data.
That is, $\xx_i \in \mathcal{D}_L \cup \mathcal{D}_U$ are first clustered after applying the feature selection $\phi$. Then, we compute \eqref{eq:f*} by replacing $\mathcal{D}_{*,p}^\phi$ with $\mathcal{D}_{L}^\phi = \{\phi(\xx_i) : (\xx_i, y_i) \in \mathcal{D}_L\}$. Therefore, our objective function becomes
\begin{equation}
f(\phi) = \max_{\mathcal{K} \in 2^{\llbracket 1, K\rrbracket}} \frac{ \abs{ \mathcal{D}_L^\phi \cap \bigcup_{k \in \mathcal{K} } \mathcal{C}_k}  }{ \abs{\mathcal{D}_L^\phi}}  \frac{ \abs{ \mathcal{D}_L^\phi \cap \bigcup_{k \in \mathcal{K} } \mathcal{C}_k}  }{ \abs{\mathcal{C}_k} } .\label{eq:f}
\end{equation}
It amounts to computing the product of the recall and the precision of the task of estimating whether the label is given or not. 

\begin{algorithm}[t]\small
\caption{Objective Function}\label{alg:objective}
\begin{algorithmic}[1]
\REQUIRE feature selector $\phi$
\REQUIRE labeled dataset $\mathcal{D}_L$, unlabeled dataset $\mathcal{D}_U$, number of clusters $K$, clustering algorithm
\STATE \texttt{\# feature extraction}
\STATE $\mathcal{D}_L^\phi = \{\phi(\xx) : \xx \in \mathcal{D}_L\}$ and $\mathcal{D}_U^\phi = \{\phi(\xx) : \xx \in \mathcal{D}_U\}$
\STATE \texttt{\# clustering}
\STATE split $\mathcal{D}_L^{\phi} \cup \mathcal{D}_U^{\phi}$ in $K$ clusters $\mathcal{C}_1, \dots, \mathcal{C}_K$ 
\STATE \texttt{\# objective function value computation}
\STATE compute $f_k = \abs{\mathcal{D}_L^\phi \cap \mathcal{C}_k} / \abs{\mathcal{C}_k}$ for $k = 1, \dots, K$
\STATE sort $\{f_k\}$ in the descending order
\STATE set $\mathcal{K} = \{1\}$ and $f_1 = \abs{\mathcal{D}_L^\phi \cap \mathcal{C}_1}^2 / \abs{\mathcal{C}_1}$
\FOR{$i = 2, \dots, K$}
\STATE compute $f_i = \abs{\mathcal{D}_L^\phi \cap \bigcup_{k=1}^{i}\mathcal{C}_k}^2 / \abs{\bigcup_{k=1}^{i}\mathcal{C}_k}$
\IF{$f_{i-1} > f_i$}
\STATE \textbf{break}
\ELSE
\STATE $\mathcal{K} \gets \mathcal{K} \cup \{i\}$
\ENDIF
\ENDFOR
\RETURN $f(\phi) = \frac{ \abs{\mathcal{D}_L^\phi \cap \bigcup_{k\in \mathcal{K}}\mathcal{C}_k}^2 }{ \abs{\mathcal{D}_L^\phi} \abs{\bigcup_{k\in \mathcal{K}}\mathcal{C}_k} }$
\end{algorithmic}
\end{algorithm}

\paragraph{Efficient computation}
To compute the objective function value \eqref{eq:f}, the best subset $\mathcal{K}$ needs to be determined among $2^K$ combinations. 
Fortunately, there is an algorithm that selects the best subset efficiently (linear in $K$). 
The procedure is described in \Cref{alg:objective}.
First, we compute the ratio of the labeled data in each cluster,
\begin{equation}
\frac{\abs{\mathcal{D}_L^\phi \cap \mathcal{C}_k}}{\abs{\mathcal{C}_k}}.
\end{equation}
Let $k_i$ be the index of the cluster with the $i$th greatest ratio. 
For $i = 1$, we compute \eqref{eq:f} (without $\max_{\mathcal{K}}$) for $\mathcal{K}_{1} = \{k_1\}$. 
For $i = 2, \dots, K$, we compute \eqref{eq:f} (without $\max_{\mathcal{K}}$) for $\mathcal{K}_{i} = \{k_1, \dots, k_i\}$ and check if the value is decreased. If \eqref{eq:f} for $\mathcal{K}_{i+1}$ is smaller than that for $\mathcal{K}_{i}$, we stop the procedure and $\mathcal{K} = \mathcal{K}_{i}$ is the best solution.
This procedure is validated by the following result, whose proof is provided in \Cref{apdx:proof} in the supplementary material.
\begin{proposition}\label{prop:valid}
Let $\mathcal{K}_{*}$ be the subset of indices that maximize \eqref{eq:f}. 
In case there is more than one optimal subset, we select the greatest subset.
Then, for any pair $(k, \ell)$ such that $k \in \mathcal{K}_*$ and $\ell \notin \mathcal{K}_*$, 
\begin{equation}
    \frac{\abs{\mathcal{D}_{L}^\phi \cap \mathcal{C}_k}}{\abs{\mathcal{C}_k}}
    > \frac{\abs{\mathcal{D}_{L}^\phi \cap \mathcal{C}_\ell}}{\abs{\mathcal{C}_\ell}}.
\end{equation}
\end{proposition}

\paragraph{Relation to $F_1$-score}
This metric is similar to the $F_1$ score, but we argue that this metric is more reasonable than the well-known $F_1$-score in the setting of PU learning. The reason is as follows. 
Let $g_{\phi, \mathcal{K}_*^\phi}$ be the discriminator as before, where $\mathcal{K}_*^\phi$ be the set of indices that maximizes \eqref{eq:f}. 
For conciseness, we write it as $g$ here.
Suppose that the data follow the distribution of $(X, Y, S)$, where $X$ and $Y$ represent the input and the label, and $S$ represents whether the label is revealed ($S = 1$) or not ($S = 0$). 
Then, \eqref{eq:f} is regarded as a sample approximation of its population version
\begin{multline}
\Pr[g(X) = 1 \mid S = 1] \Pr[S = 1 \mid g(X) = 1] \\
= \frac{\Pr[g(X) = 1 \wedge S = 1]^2}{ \Pr[S = 1] \Pr[g(X) = 1]}.\label{eq:pop-f}
\end{multline}
In the PU learning setting, it is often assumed that the labels are missing completely at random (MCAR assumption), that is, $\Pr[S = 1 \mid X, Y=1] = \Pr[S=1\mid Y=1] = \beta$ for some constant $\beta \in (0, 1]$. 
Under the MCAR assumption as well as the assumption of the conditional independence of $g(X)$ and $S$ conditioned on $Y$ (this is a natural assumption as $g(X)$ is obtained without using labels), we have $\Pr[S = 1 \wedge g(X) = 1] = \Pr[S = 1 \mid Y = 1] \Pr[g(X) = 1 \mid Y = 1] = \beta \cdot \Pr[g(X) = 1 \mid Y = 1]$ and $\Pr[S = 1] = \Pr[S = 1 \mid Y = 1] \Pr[Y = 1] = \beta \cdot \Pr[Y = 1]$.
Then, we obtain
\begin{equation}
\mathrm{Eq.}~\eqref{eq:pop-f}
= \beta \frac{\Pr[g(X) = 1 \mid Y = 1]^2}{ \Pr[Y = 1] \Pr[g(X) = 1]}.
\end{equation}
It indicates that the discriminator $g$ maximizing this metric is invariant to the change of the label drop rate $1 - \beta \in (0, 1]$. In particular, the optimal $g$ for the PU dataset is optimal for the fully labeled dataset in terms of the product of the recall and the precision.
On the other hand, the $F_1$-score is not invariant under the same assumption as
\begin{equation}
\begin{split}
F_1\text{-score} &=2\frac{\Pr[g(X) = 1 \mid S = 1] \Pr[S = 1 \mid g(X) = 1]}{\Pr[g(X) = 1 \mid S = 1] + \Pr[S = 1 \mid g(X) = 1]}
\\
&= 2\frac{\beta \Pr[g(X) = 1 \mid Y = 1]}{\beta \Pr[Y = 1] + \Pr[g(X) = 1]},
\end{split}
\end{equation}
whose optimal $g$ changes for varying $\beta$.

\paragraph{Theoretical Insight}
To gain further insight into the proposed objective function, 
we investigate the property of $\mathcal{K}_*$ maximizing \eqref{eq:f}.
In particular, we are interested to know whether $\mathcal{K}_*$ maximizing \eqref{eq:f} is the optimal subset of clusters that corresponds to positive data. 
\Cref{prop:mcar} states that under the MCAR assumption, the subset $\mathcal{K}_*$ maximizing \eqref{eq:f} indeed corresponds to the set of positive clusters if the data are adequately clustered.
Its proof is provided in \Cref{apdx:proof} in the supplementary material.
\begin{proposition}\label{prop:mcar}
Suppose that $\abs{\mathcal{C}_i} = n_i$ for $i = 1, \dots, K$ and $\frac{\abs{\mathcal{D}_{L}^\phi \cap \mathcal{C}_i}}{\abs{\mathcal{C}_i}} = \beta$ for $i = 1, \dots, K_A$ and $\frac{\abs{\mathcal{D}_{L}^\phi \cap \mathcal{C}_i}}{\abs{\mathcal{C}_i}} = 0$ for $i = K_A + 1, \dots, K$, for some $K_A \in \llbracket 1, K\rrbracket$. Then, $\mathcal{K}_*$ maximizing \eqref{eq:f} is $\{1, \dots, K_A\}$.
\end{proposition}
However, if the MCAR assumption is not satisfied, $\mathcal{K}_*$ may not correspond to the set of positive clusters even if the data are adequately clustered. We leave further investigation into this direction for future work.

\subsection{Compact-GA-based optimizer}\label{sec:cga}

We employ an evolutionary approach to optimize the proposed objective function for feature selection in the PU learning setting. 
In particular, we employ compact genetic algorithms \cite{cga}. 
It is also known as population-based incremental learning \cite{pbil} and an instantiation of information-geometric optimization \cite{igo} for optimization of binary variables.
We introduce a modification to the original algorithm to address the task-specific requirement.

We formulate the problem of feature selection as the optimization of binary variables. 
Let $m \in \mathcal{M} = \{0, 1\}^d$ be a binary vector, each element of which determines whether the corresponding feature is used ($m_i = 1$) or not ($m_i = 0$), where $d$ is the number of features.
That is, $m$ determines $\phi$. 
Our objective is to find the mask pattern $m^\star$ that maximizes the objective function $f(m)$, where $f(m)$ represents the objective function value $f(\phi)$ defined in \eqref{eq:f} with $\phi$ determined by $m$ with a slight abuse of notation, under the constraint that the selected features must satisfy the cost constraint, $\sum_{\ell=1}^{d} m_\ell c_\ell \leq \bar{c}$, where $c_\ell$ is the cost of selecting the $\ell$-th feature and $\bar{c}$ is the total cost allowed. A typical situation where the constraint is required is that we would like to select at most some predefined number of features. In this case, we can set $c_\ell = 1$ for all $\ell$ and the threshold $\bar{c}$ is the upper bound of the number of features to be selected.

\Cref{alg:cga} summarizes our search algorithm.
For simplicity, we first introduce the algorithm for an unconstrained problem. 
It maintains Bernoulli distribution with parameter $\theta \in \Theta = [\epsilon, 1-\epsilon]^d$, from which candidate feature selection patterns are sampled for search, where $\theta_\ell$ represents the probability of selecting the $\ell$-th feature ($m_\ell = 1$), and $\epsilon$ is the clip rate.
After the initialization of $\theta$ (described later), two samples, $m^a$ and $m^b$, are drawn from the Bernoulli distribution. They are evaluated on the objective function, $f(m^a)$ and $f(m^b)$.
Then, the parameter $\theta$ is updated so that a better candidate becomes more likely to be sampled as
\begin{align}
\theta \gets \theta + \eta \cdot \textsc{sign}(f(m^a) - f(m^b)) (m^a - m^b)  ,\label{eq:update}
\end{align}
where $\eta \in [0, 1]$  is the learning rate. $\theta$ is always forced to be in $\Theta$ by simply performing
\begin{align}
\theta_i \gets \min(\max(\theta_i, \epsilon), 1 - \epsilon).
\end{align}
This process is repeated for a given number of iterations, $T$, or until some other termination criterion is satisfied.
The update \eqref{eq:update} can be viewed as the natural gradient ascent of the expected objective function $\E_{m \sim p_\theta}[ f(m) ]$. 
That is, this algorithm tries to maximize the expected objective function by following the natural gradient with respect to the parameter vector, which results in minimizing the objective function.
See \cite{igo} for details.

\begin{algorithm}[t]\small
\caption{IGO with Hard Constraint}\label{alg:cga}
\begin{algorithmic}[1]
\STATE $\theta^0 =  (\theta_0, \dots, \theta_0)$, where $\theta_0 = \bar{c} / \sum_{\ell} c_{\ell}$
\FOR{$t = 1, \dots, T$}
\STATE $m^a, m^b \sim p_\theta$ 
\STATE $\tilde{m}^a, \tilde{m}^b \gets \textsc{Repair}(m^a), \textsc{Repair}(m^b)$ 
\STATE $\theta^{t} \gets \theta^{t-1} + \eta_\theta \textsc{Sign}(f(\tilde{m}^a)-f(\tilde{m}^b)) (\tilde{m}^a - \tilde{m}^b)$
\STATE $\theta^{(t)} \gets \texttt{clip}(\theta^{(t)}, \epsilon, 1-\epsilon)$
\ENDFOR
\end{algorithmic}
\end{algorithm}

\begin{algorithm}[t]\small
\caption{\textsc{Repair}}\label{alg:repair}
\begin{algorithmic}[1]
    \REQUIRE $m$
    \STATE $\tilde{m} \gets m$
    \STATE \texttt{\# to satisfy the constraint}
    \WHILE{$\sum_{\ell} \tilde{m}_{\ell} c_{\ell} > \bar{c}$}
    \STATE $\ell' \sim \textsc{Categorical}\left(p = \frac{\tilde{m}_{\ell}(1 -  \theta_{\ell})}{\sum_{\ell^*} \tilde{m}_{\ell^*}(1 -  \theta_{\ell^*})}\right)$
    \STATE $\tilde{m}_{\ell'} = 0$
    \ENDWHILE
    \STATE \texttt{\# to get closer to $\bar{c}$}
    \STATE $\tilde{c} = \bar{c} - \sum_{\ell} \tilde{m}_{\ell} c_{\ell}$
    \WHILE{$\tilde{c} \geq \min_{\ell: \tilde{m}_{\ell} = 0} c_{\ell}$}
    \STATE $\ell' \sim \textsc{Categorical}\left(p = \frac{ \ind{c_{\ell} \leq \tilde{c}} (1 - \tilde{m}_{\ell}) \theta_{\ell} }{ \sum_{\ell^*} \ind{c_{\ell} \leq \tilde{c}} (1 - \tilde{m}_{\ell}) \theta_{\ell} } \right)$
    \STATE $\tilde{m}_{\ell'} = 1$
    \STATE $\tilde{c} = \bar{c} - \sum_{\ell} \tilde{m}_{\ell} c_{\ell}$
    \ENDWHILE
    \RETURN $\tilde{m}$
\end{algorithmic}
\end{algorithm}

Now we discuss how we treat the cost constraint $\sum_{\ell=1}^{d} m_\ell c_\ell \leq \bar{c}$.
The motivation for the cost constraint is twofold. 
First, our objective is to select a subset of features that perform well on downstream tasks and that are interpretable for later data analysis. 
%
From this perspective, the cost constraint only needs to be satisfied by the final solution, while intermediate solutions during optimization can violate it. 
Second, we would like to avoid evaluating the objective function value for $m$ violating the cost constraint because the computational cost for the objective function value increases as the number of $1$-bit in $m$ increases. From this point of view, we would like to force the candidate solutions to satisfy the cost constraints. 
Therefore, we propose a repair operator to satisfy the constraint. 
The proposed repair operator is described in \Cref{alg:repair}.
If a candidate solution $m$ violates the constraint, one bit of $m$ is selected and is flipped to $0$. 
The bit is selected based on the parameter value $\theta_\ell$. 
If $\theta_\ell$ is smaller, it is more likely to be flipped. 

Our repair operator also implements the functionality to make the cost closer to the upper bound.
The motivation for the introduction of this functionality is the observation that the performance of the downstream tasks after feature selection is typically better when more features are selected. 
Therefore, we want to select as many features as possible, while satisfying the cost constraint.
For this purpose, if a given candidate solution $m$ does not reach the cost threshold, we select one bit of $m$  and flip to $1$. 
The bit is selected based on the parameter value $\theta_\ell$.

\section{Experiments with artificial data}\label{ExperimentsArtificialData}
We conducted experiments using artificial data to identify the data conditions under which FSCPU performs well. 

\subsection{Hypothesis}\label{Hypothesis}
In these experiments, we confirm the following hypothesis: FSCPU improves performance under the following conditions.
\begin{enumerate}
\item \label{Hypothesis:first} The peaks in the distribution of positively labeled data exhibit agglomerative properties.
\item \label{Hypothesis:second} The number of labeled datapoints is small.
\item \label{Hypothesis:third} The distribution of positively labeled data exhibits multimodality.
\item \label{Hypothesis:fourth} The clustering process is not dominated by a large proportion of data that should not be labeled positive, ensuring the appropriate classification of the true positive data distribution.
\end{enumerate}

Condition~(\ref{Hypothesis:first}) reflects the constraints of the clustering assumption, which forms the foundation of FSCPU. 
Condition~(\ref{Hypothesis:second}) suggests that, as the number of labeled data points increases, FSCPU may become less effective compared to other approaches that directly leverage the statistical properties of labeled data.
Condition~(\ref{Hypothesis:third}) assumes that FSCPU remains robust even when handling multimodal positive data distributions.
Condition~(\ref{Hypothesis:fourth}) reflects the constraints associated with the use of the clustering model.
In addition to the aforementioned constraints, FSCPU would not function effectively when using distance-based clustering unless the input data contain features that are relevant to the chosen distance metric.

\subsection{Experimental Settings}

\subsubsection{Artificial Data}\label{ArtificialData}
The entire dataset consists of 50 features; of these, 25 are relevant features, while the remaining 25 are irrelevant to the labels.
Among the 25 irrelevant features, 20 were sampled from a uniform distribution in the range of -10 to 10.
The remaining 5 irrelevant features were generated by randomly selecting 5 features from the aforementioned 20 and adding noise. 
The noise was sampled from a normal distribution with a mean of 0 and a variance of 1.
This was intended to ensure that some of the irrelevant features had a certain degree of correlation.
Datasets were generated under different conditions for data that should receive a positive label and data that should receive a negative label.
The PU learning setting was simulated by labeling only a portion of the positive labels in the dataset.
Two variations of the labeling rate, $40\%$ and $10\%$, were established as situations of more and less labeling, respectively.
In the experiments, the data points generated by each normal distribution were considered to form a cluster.
We created 8 artificial datasets satisfying the cluster assumption by generating positive-labeled data points from a normal distribution with a unique mean.
For negative example data, 4,000 rows and 25 columns were sampled from a normal distribution of one or eight different means.
Positive example data consisted of 500 rows and 25 columns sampled from a normal distribution of one or two different means.
The mean of each normal distribution was sampled randomly between -5 and 5, with a variance of 10.
We created two artificial datasets that do not satisfy the cluster assumption by generating positively labeled data as scattered outliers from a normal distribution with a specific mean. 
First, a dataset consisting of 4,500 rows and 25 columns was sampled from a normal distribution with a mean of 0 and a variance of 25. 
The 500 rows with the highest norms were then assigned positive labels. 
This corresponds to labeling data that are more than +1.22$\sigma$ from the mean as positive.

\begin{table*}[t]\small
    \caption{Results on artificial data. 
    The mean and standard deviation of FSR are shown. 
    ``cluster assumption'': whether the cluster assumption is satisfied or not;
    ``labeled rate'': the proportion of the positive data for which the labels are given;
    ``no.\ negative cluster'' and ``no.\ positive cluster'': the numbers of normal distributions associated to negative clusters and positive clusters.
}
    \label{tab1}
    \begin{center}
    \begin{tabular}{ccccccccccccc}
    \toprule
        \multicolumn{2}{c}{cluster assumption} & $\checkmark$ & $\checkmark$ & $\checkmark$ & $\checkmark$ & $\checkmark$ & $\checkmark$ & $\checkmark$ & $\checkmark$ & $\times$& $\times$\\ \midrule
        \multicolumn{2}{c}{labeled rate} & $40\%$& $40\%$& $40\%$& $40\%$ & $10\%$& $10\%$& $10\%$& $10\%$& $40\%$& $10\%$\\ \midrule
        \multicolumn{2}{c}{no.\ negative cluster} & $8$& $8$ & $1$& $1$& $8$& $8$ & $1$& $1$& -& -\\ \midrule
        \multicolumn{2}{c}{no.\  positive cluster} & $1$ & $2$& $1$& $2$& $1$ & $2$& $1$& $2$& -& - \\ \midrule
        \multirow{5}{*}{supervised}&K-Best &.82±.10&.78±.05&.87±.04&.86±.08&.67±.08&.66±.06&.79±.07&.76±.06&.42±.05&.35±.03 \\ 
        &Lasso  &.86±.05&.86±.05&.83±.03&.84±.05&.78±.07&.78±.02&.75±.05&.76±.06&.66±.07&.62±.05 \\ 
        &mRMR   & .10±.02&.12±.03&.17±.04&.19±.08&.10±.02&.10±.02&.10±.04&.10±.05&.26±.09&.11±.04 \\ 
        &DT-RFE    & .74±.07&.83±.06&.78±.07&.80±.04&.54±.05&.59±.07&.62±.07&.63±.03&1.0±.00&.89±.03 \\ 
        &ES-MCRFS & .43±.16&.49±.08&.56±.09&.44±.06&.50±.05&.50±.10&.48±.03&.54±.05&.41±.17&.52±.10 \\ \midrule
        unsupervised&MRMR-UFS & .59±.02&.59±.03&.61±.02&.60±.00&.59±.02&.59±.03&.61±.02&.60±.00&.60±.00&.60±.00 \\ \midrule
        \multirow{2}{*}{semi-supervised}&sSelect & .34±.06&.36±.04&.42±.07&.45±.07&.34±.06&.35±.05&.42±.07&.45±.07&.35±.10&.35±.10 \\ 
        &JSFS & .43±.07&.51±.03&.42±.05&.49±.05&.46±.10&.44±.05&.42±.05&.49±.03&.42±.07&.46±.07 \\ \midrule
        \multirow{3}{*}{\shortstack{positive-unlabeled\\(semi-supervised)}}&semi-JMI & .97±.03&.98±.02&.89±.05&.94±.05&.70±.12&.69±.08&.71±.10&.79±.03&.93±.02&.57±.08 \\ 
        &FSCPU(ours) &  .91±.07&.90±.08&.75±.09&.82±.06&.92±.06&.89±.07&.78±.09&.78±.07&.56±.09&.51±.10 \\ 
        &FSCPU-MI(ours) & .90±.06&.79±.09&.74±.06&.82±.06&.87±.06&.86±.06&.76±.14&.81±.03&.59±.10&.54±.08 \\ 
        \bottomrule  
    \end{tabular}
    
    \end{center}
\end{table*}

\subsubsection{Baseline}\label{Baseline}
We consider the following baseline approaches.

(1) \textbf{K-Best} algorithm \cite{K-best_b5} computes a $\chi^2$ score between the label vectors and features, selecting the features with the highest scores. 
In this study, K-Best was implemented using \texttt{SelectKBest} and \texttt{chi2} functions 
from \texttt{scikit-learn} library with default parameters. 

(2) \textbf{Lasso} performs feature selection using an $\ell_1$ penalty \cite{LASSO_b6}. 
In our experiments, Lasso was implemented using \texttt{Lasso} function from \texttt{scikit-learn} library, with the regularization parameter \texttt{alpha} set to $5 \times 10^{-5}$. 
Feature selection was performed by selecting the specified number of features in descending order of the absolute values of the learned weights.

(3) \textbf{DT-RFE} \cite{DT-RFE_b7} is a method for training a decision tree on all the features and determining their importance. 
Features of low importance were deselected by repeating the process  
until a set number of selections was reached. 
DT-RFE was implemented by combining \texttt{RFE} and \texttt{RandomForestClassifier} from \texttt{scikit-learn} library. 
The random forest classifier was set to use $100$ decision trees.

(4) \textbf{mRMR} \cite{mRMR_b8} derives a score for each feature based on an evaluation index that reduces the mutual information between the features and increases the mutual information between the label vector and features. 
Features with high scores are selected. 
The mRMR algorithm was implemented using  \texttt{pymrmr} library, employing \texttt{MIQ} method with default parameters.

(5) \textbf{MRMR-UFS} \cite{MRMR-UFS_b10} is an unsupervised feature-selection method based on spectral clustering.
Maximum Relevance Minimum Redundancy (MRMR) is considered.
In this study, we used MRMR-UFS2 \cite{MRMR-UFS_b10}, which allows the number of selections to be specified and is considered to have good performance.
In the experiments, the 5-nearest neighbor method was used. The number of clusters for features, denoted as $K_1$ and $K_2$ in \cite{MRMR-UFS_b10}, was set equal to the number of selected features.

(6) \textbf{sSelect} \cite{sSelect_b4} is a semisupervised feature selection method that uses the spectral graph theory and cluster assumption. 
A cluster index is computed using Laplacian scores, with the goal that all labeled data in each cluster come from the same class. 
In the experiment, the regularization parameter $\lambda$ was set to 0.1, and a neighborhood graph was constructed with a neighborhood size of 5.

(7) \textbf{JSFS} \cite{JSFS_b3} is a Bayesian approach for simultaneously updating the feature selection parameters and unlabeled sample weights.
Through iteration, the amount of data to be computed decreases, and the final computational cost decreases.
In the experiment, the number of nearest neighbors was set to 5, the parameter $\gamma$ was set to $1 \times 10^{-11}$, and $\mu$ was set to $9.95 \times 10^{-3}$.

(8) \textbf{Early stopping Monte Carlo-based reinforced feature selection (ES-MCRFS)} \cite{ES-MCRFS_single_agent_b9} is a feature selection method based on single-agent reinforcement where a single agent sequentially controls the selection or deselection of all feature sets. In addition, an early stopping strategy that terminates the search for features estimated to be ineffective allows for efficient feature selection.
ES-MCRFS was implemented with a mini-batch size of 8 for deep network training, utilizing the Adam optimizer with a learning rate of $0.001$. 
The mean squared error (MSE) loss function was employed, with $\tanh$ activation functions for the hidden layers and a sigmoid activation function for the output layer.
The number of search steps was set to $6,000$, and the final result was determined based on the last output that satisfied the specified number of selected features.
All other parameters were set according to the original paper \cite{ES-MCRFS_single_agent_b9}.

(9) \textbf{semi-JMI} \cite{semi_JMI_b21} is a feature selection method for PU learning settings that utilizes mutual information. It identifies the feature selection pattern with the highest mutual information content within a predefined number of features by iteratively adding and removing features from the label data and candidate features.
Semi-JMI was implemented based on the literature \cite{semi_JMI_b21}. 

\subsubsection{FSCPU with Mutual information criterion (FSCPU-MI)}\label{sec:mi}
As a variant to FSCPU, we consider combining the proposed objective function $f(\phi)$ defined in \eqref{eq:f} with a mutual information criterion $\hat{I}(\phi)$ employed in \cite{semi_JMI_b21}. 
The combined objective function is expressed by the following equation:
\begin{equation}
   \frac{f(\phi)}{Std[f]}  + \frac{\hat{I}(\phi)}{Std[\hat{I}]}.
    \label{eq:55}
\end{equation}
The scores $(f(\phi), \hat{I}(\phi))$ are obtained in each iteration and appended to their respective score logs (unbounded). The standard deviations, $Std[f]$ and $Str[\hat{I}]$, of the score logs are then calculated, and the coefficients are computed based on these standard deviations.

\subsubsection{Implementation of the proposed methods.}\label{ProposedMethodsSettings}
In FSCPU, the objective function described in Section \ref{sec:obj} was optimized using the optimizer presented in Section \ref{sec:cga}.
We set $\eta_\theta = 1/2d$, $\epsilon = 1/d$, $c_\ell = 1$, $T=3,000$, and $\bar{c}$ as the number of features to be selected throughout our experiments.
A Gaussian Mixture Model (GMM) was adopted for clustering, by using 10 GMM kernels utilized ($K=10$). 
The variant where the objective function described in Section \ref{sec:mi} was used is referred to as FSCPU-MI. 
After the exploration process, the features corresponding to the top $\theta$ values were selected based on the specified number of features to be chosen.

\subsubsection{Evaluation metrics}\label{ArtificialDataEvaluations}
Experiments were evaluated on the basis of the feature selection recall (FSR) of the artificial data. 
The FSR is defined as the number of correctly selected relevant features 
divided by the total number of relevant features: 
\begin{equation}
    \text{FSR} = \frac{\text{Number of correctly selected relevant features}}{\text{Total number of relevant features}}.
\end{equation}

\subsubsection{Experimental Procedure}\label{ArtificialDataOtherSettings}
First, the feature values in the artificial dataset were scaled using the min-max normalization method.
After preprocessing, the artificial data were input into the feature selection methods.
In each experiment, 25 features were selected.
Feature selection performance was evaluated based on the FSR, which equals 1.0 if all relevant features were correctly selected.
Each method was evaluated by conducting five experiments on each dataset.

\begin{figure*}
    \centering
    \begin{subfigure}[t]{0.25\textwidth} 
        \centering
        \includegraphics[width=\linewidth]{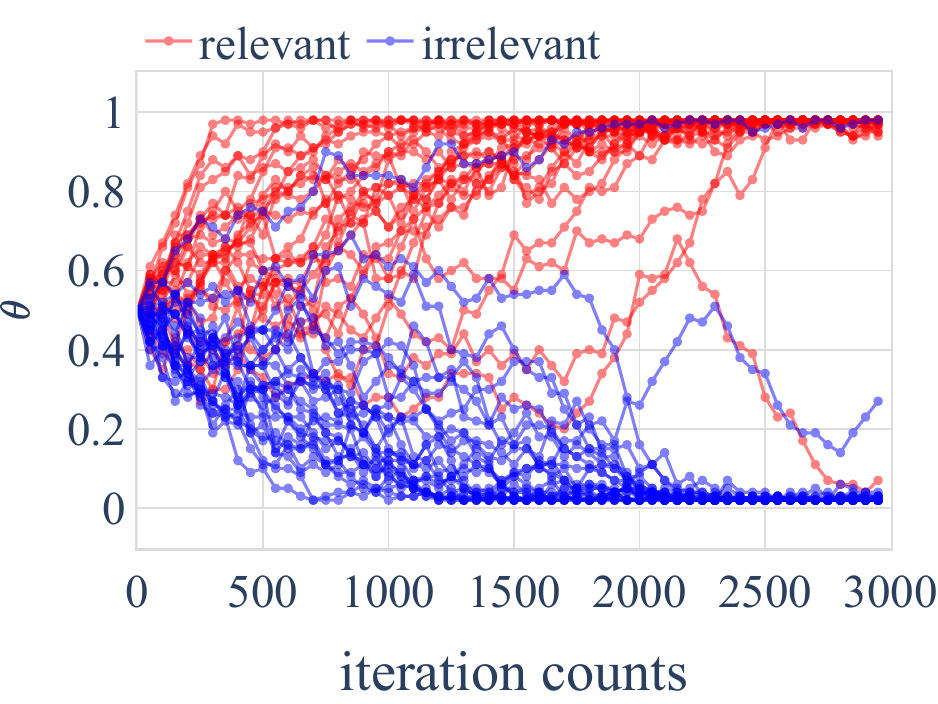}
        \caption{conditions = \{$\checkmark$, 40\%, 8, 2\}} 
    \end{subfigure}%
    \begin{subfigure}[t]{0.25\textwidth}
        \centering
        \includegraphics[width=\linewidth]{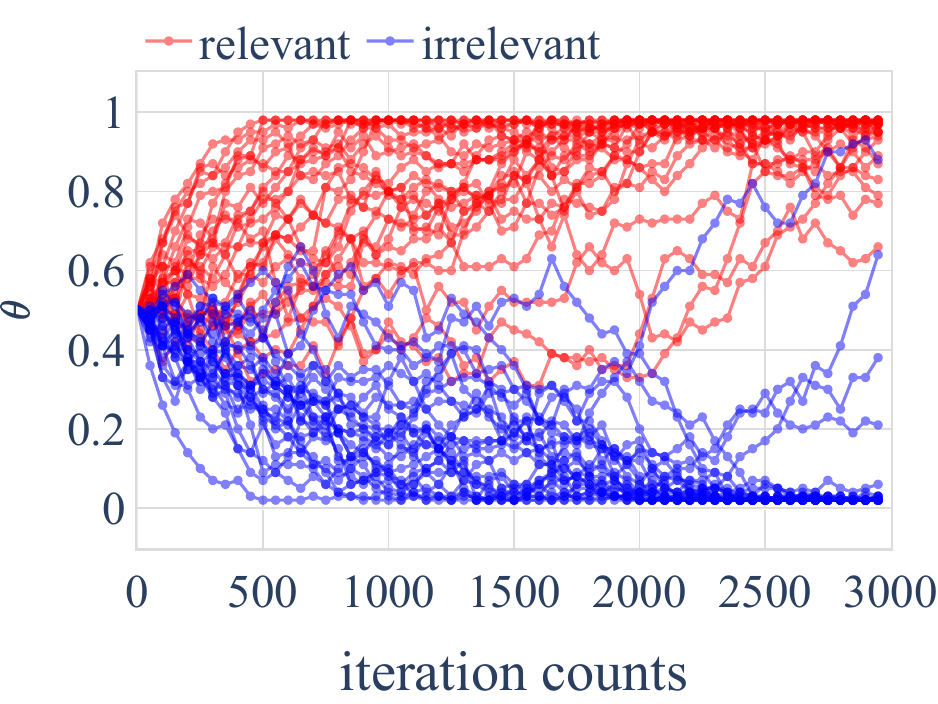}
        \caption{conditions = \{$\checkmark$, 10\%, 8, 2\}}
    \end{subfigure}%
    \begin{subfigure}[t]{0.25\textwidth}
        \centering
        \includegraphics[width=\linewidth]{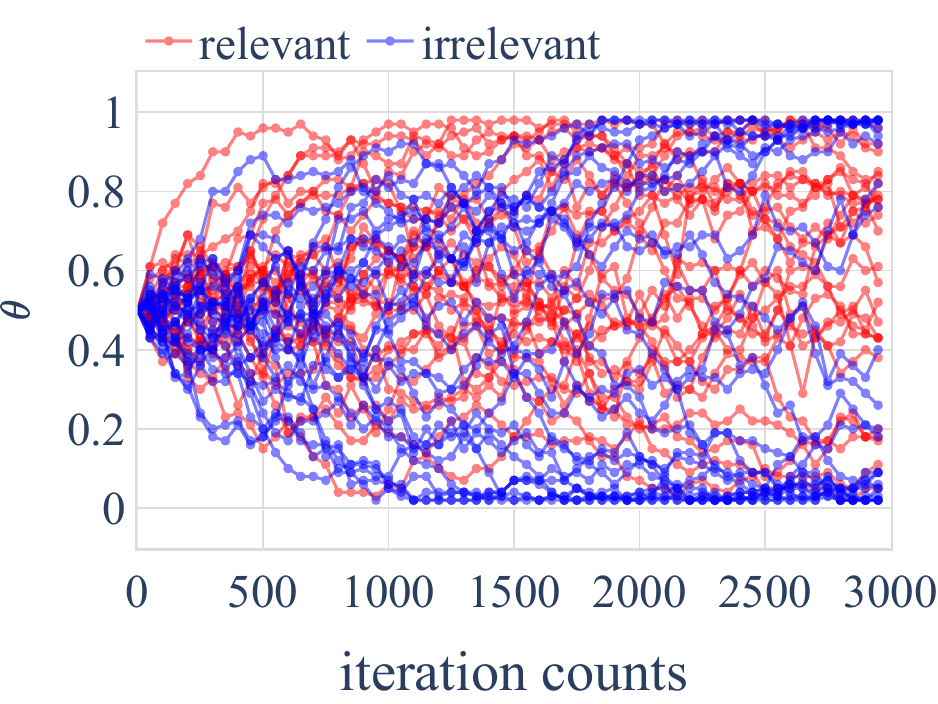}
        \caption{conditions = \{$\times$, 10\%\}}
    \end{subfigure}%
    \begin{subfigure}[t]{0.25\textwidth}
        \centering
        \includegraphics[width=\linewidth]{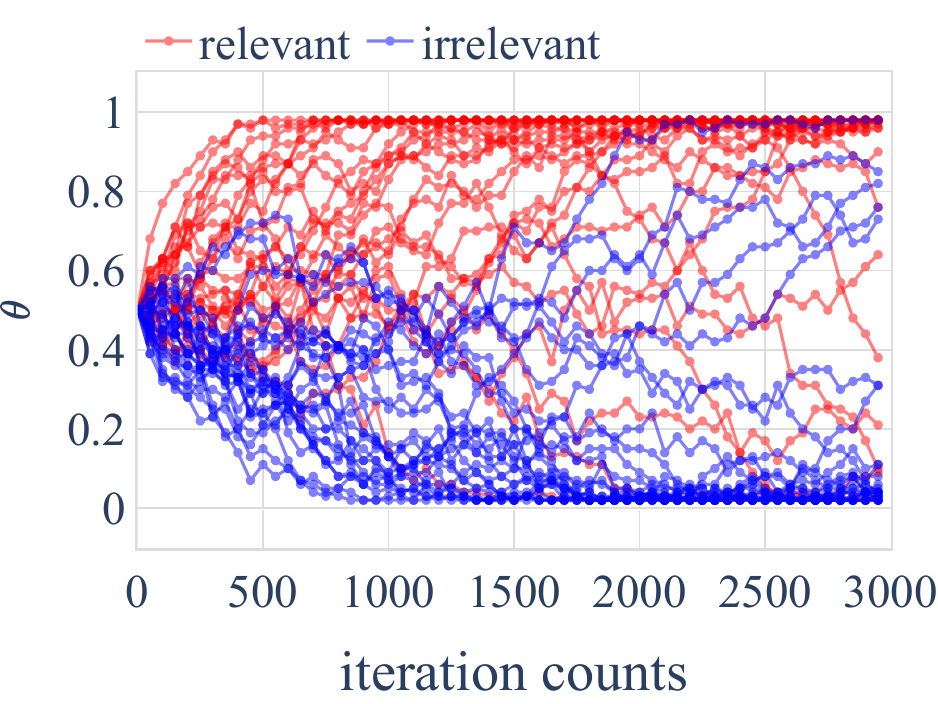}
        \caption{conditions = \{$\checkmark$, 10\%, 1, 2\}}
    \end{subfigure}%
    \caption{Convergence behavior of probability parameter $\theta$ on different datasets. 
    Red and blue lines indicate $\theta_i$ corresponding to relevant and irrelevant features, respectively.
    The subfigure captions describe different experimental conditions corresponding to the top four rows of Table~\ref{tab1} (i.e., conditions = \{cluster assumption, labeled rate, no. negative cluster, no. positive cluster \}).}
    \label{fig:example}
\end{figure*}

\subsection{Results}\label{ArtificialDataResults}
Table \ref{tab1} provides insights into the validity of the hypotheses presented in Section~\ref{Hypothesis}.
The overall performance of the proposed methods (FSCPU and FSCPU-MI) was higher in experimental patterns that satisfy the cluster assumption, thereby confirming the validity of the hypothesis regarding Condition~(\ref{Hypothesis:first}).
With the datasets satisfying the cluster assumption, when the labeling rate decreases from $40\%$ to $10\%$, most conventional methods either experienced a decline in their scores or their scores remained low, whereas the proposed methods maintained a relatively high score, thus confirming the validity of Condition~(\ref{Hypothesis:second}).
The scores of FSCPU did not significantly change when the number of positive clusters from 1 to 2, confirming the validity of Condition~(\ref{Hypothesis:third}). However, because many of the other methods also exhibited relatively stable scores under these conditions, this scenario may not present a unique advantage for the FSCPU.
Finally, when comparing the results for data satisfying the cluster assumption, a decrease in the score of FSCPU when the number of negative clusters decreases from 8 to 1 confirms the validity of Condition~(\ref{Hypothesis:fourth}).
The decrease occurs because concentrating a large volume of data points into a single negative distribution causes fragmentation during clustering, hindering the appropriate separation of the true positive data distribution.

Figure~\ref{fig:example} shows results that support the above considerations in terms of $\theta$ convergence.
First, when the data conditions of Section~\ref{Hypothesis} are satisfied (Figure~\ref{fig:example}(b)), 
the convergence is close to that of the case with many labels (Figure~\ref{fig:example}(a)).
Furthermore, it was observed that when the cluster assumption is not met (Figure~\ref{fig:example}(c)), 
convergence is poor, and when the unlabeled cluster number is 1 (Figure~\ref{fig:example}(d)), convergence is also relatively poor.

\section{Experiments with open data}
To examine the effectiveness of FSCPU in a near-realistic setting, experiments were conducted using open datasets. 
In this study, the feature selection results were evaluated based on the performance of downstream classification models, a common approach for assessing the effectiveness of filter methods.
\subsection{Hypothesis} \label{OpenDataHypothesis}
In this section, we intend to confirm the following hypothesis: FSCPU can achieve performance comparable to that of conventional methods, even when only the condition of having a small number of labels is satisfied.

This hypothesis argues that FSCPU is useful even if it only satisfies Condition (\ref{Hypothesis:second}) discussed in Section \ref{Hypothesis}.
The conditions in Section \ref{Hypothesis} other than Condition (\ref{Hypothesis:second}) are challenging to control or verify in advance when using real-world data.
The results in Table \ref{tab1}, with a labeling rate of 10\%, suggest that FSCPU achieved performance comparable to conventional methods only under Condition (\ref{Hypothesis:second}).
Based on this suggestion, we focus on the hypothesis.

\subsection{Experimental Settings}\label{CommonSetting}

\subsubsection{Dataset}
Table \ref{tab2} summarizes the datasets used for the experiments.

(1) \textbf{Spambase (Spam) dataset}: 
The original dataset \cite{spambase_94} contained binary labels of 0 and 1, with 1,813 records labeled as spam (1) and 2,788 records labeled as non-spam (0). 
The total of 4,601 records included 57 features. 
In the experiments, positive labels were assigned to a randomly selected 3\% of the spam (1) data.

(2) \textbf{Ionosphere (Iono) dataset}: 
The original dataset \cite{ionosphere_52} contained binary labels, with 126 records labeled as bad (1) and 225 records labeled as good (0). 
The dataset consisted of a total of 351 records with 34 features. 
In the experiments, positive labels were assigned to a randomly selected 10\% of the bad (1) records.

(3) \textbf{RT-IoT2022 (RT-IoT) dataset}: 
The original dataset \cite{rt-iot2022__942} consists of 12 different labels, 123,117 records, and 81 features. 
In this study, 67 features were selected for the experiments, excluding categorical features. 
Furthermore, data with two types of labels, \texttt{NMAP\_TCP\_scan} and \texttt{DDOS\_Slowloris}, were adopted as positive data.
One label, \texttt{DOS\_SYN\_Hping}, was excluded owing to an excessively high number of assigned instances. 
In the experiments, positive labels were assigned to a randomly selected 2\% of the positive data.

\begin{table}[t]\small
    \caption{Overview of open datasets. 
    $\text{``p-instances''}$ refers to positively labeled instances in the dataset.}
    \begin{center}
    \begin{tabular}{cccc}
    \toprule
        Dataset & Spam & Iono & RT-IoT \\ \midrule
        the number of classes& 2 & 2 & 11 \\
        the number of instances& 4,601 & 351 & 28,458 \\
        the number of features & 57 & 34 & 67 \\
        the total number of p-instances & 1,813 & 126 & 1,536 \\
        the number of p-instances for training & 40 & 9 & 23 \\
        the number of p-instances for testing & 453 & 31 & 384 \\
        \bottomrule  
    \end{tabular}
    \end{center}
    \label{tab2}
\end{table}

\begin{table}\small
    \caption{Results on open datasets.
    The mean and standard deviation of AUC are shown. 
    The ``labeled rate'' indicates the proportion of positive data for which the labels are provided.
    }
    \label{tab3}
    \begin{tabular}{ccccc}
    \toprule
        \multicolumn{2}{c}{Dataset} & Spam & Iono & RT-IoT \\ \midrule
        \multicolumn{2}{c}{labeled rate} & $3\%$ & $10\%$ & $2\%$ \\ \midrule
        \multirow{6}{*}{supervised}&K-Best & .866±.011&.832±.030&.945±.018 \\ 
        &Lasso  & .869±.019&.834±.040&.983±.002 \\ 
        &mRMR   & .809±.011&.864±.036&.898±.022 \\ 
        &DT-RFE & .841±.017&.787±.068&.921±.010  \\ 
        &ES-MCRFS & .671±.037&.779±.050&.894±.017 \\ 
        &LGBM-OPT & .856±.040&.822±.041&.946±.018 \\ \midrule
        unsupervised&MRMR-UFS & .840±.035&.847±.041&.873±.013 \\ \midrule
        \multirow{2}{*}{semi-supervised}&sSelect & .839±.031&.829±.026&.911±.035 \\ 
        &JSFS & .855±.013&.825±.053&.915±.026 \\ \midrule
        \multirow{3}{*}{\shortstack{positive-unlabeled\\(semi-supervised)}}&semi-JMI & .757±.096&.821±.022&.922±.019 \\ 
        &FSCPU & .844±.019&.797±.091&.910±.017 \\ 
        &FSCPU-MI & .874±.013&.749±.070&.916±.020  \\ 
        \bottomrule  
    \end{tabular}
\end{table}

\subsubsection{Baseline}\label{OpendataBaseline}
In the experiments with open data, the following baseline was used in addition to the baselines (1)-(9) employed in the artificial data experiments: 
(10) \textbf{LGBM-OPT} was prepared for an ablation study of FSCPU in scenarios where a downstream classification model is specified.
LGBM-OPT employed the LightGBM \cite{LightGBM_b22} area under the receiver operating characteristic (ROC) curve (AUC) score 
as an evaluation metric in the optimization process. 
The LightGBM used for this metric was used as the classifier with 100 decision trees. 
This model was configured with the same parameter settings as the downstream classification model (described later). 
LGBM-OPT used $75\%$ of the training data for learning and the remaining $25\%$ for evaluation. 
Additionally, the optimization method was the same as FSCPU, including its implementation and parameter settings. 
After the exploration process, the features corresponding to the top $\theta$ values were selected based on the specified number of features to be chosen.

\subsubsection{Experimental Procedure} 
The experimental procedure comprises preprocessing, feature selection, and downstream classification, as described below.
First, the dataset was split into 75\% for training and 25\% for testing. Feature values were preprocessed using the min-max normalization method based on the training data. 
The training data consisted of partially labeled positive samples (1) and unlabeled samples (0), 
while the test data contained all originally labeled positive samples (1) and the remaining samples labeled as (0).
After preprocessing, the training data were input into the feature selection methods, and feature selection was performed on the training set. 
The number of selected features was set to half of the total number of features in each dataset, rounded up if necessary. 
Specifically, 34, 29, and 17 features were selected for RT-IoT, Spam, and Iono, respectively. 
For methods that output feature-wise scores, features were selected in descending order of score until the specified number was reached.
The downstream classification model was trained using the selected features from the training data. 
LightGBM was used as the classifier with 100 decision trees. 
The performance of the trained model was evaluated on the test data on the basis of the AUC score.
Three experiments were conducted for each method on each dataset.

\subsection{Results}
Table~\ref{tab3} summarizes the results, providing insights into the hypothesis discussed in Section~\ref{OpenDataHypothesis}. 
Unlike the results obtained with artificial data, the proposed methods did not show a clear advantage. 
Combining the MI score into FSCPU occasionally contributes to improving the performance.
On the other hand, although FSCPU exhibited considerable variability across datasets, it achieved competitive performance with existing methods in two out of the three datasets. 
Thus, the hypothesis in Section \ref{OpenDataHypothesis} holds to a certain extent.
Additionally, examining the parameters after updating FSCPU indicates potential for further performance improvement by increasing the number of iterations.
In these experiments, 40.1\% of the probability parameters, $\theta$, in FSCPU were not converged after the search process, where parameters in $[0, 0.1] \cup [0.9, 1]$ were regarded converged. 
In particular, on RT-IoT, where the proposed approach was least competitive, 78.5\% of the probability parameters were not yet converged.

\section{Conclusion}
We proposed a new optimization framework for feature selection in the PU learning setting, called FSCPU, and empirically evaluated it on artificial and open data.
First, for the optimization framework, we proposed the cluster assumption in feature selection for PU learning and designed an objective function and optimization method based on it.
To the best of our knowledge, a cluster assumption in feature selection for PU learning and 
its corresponding objective function has not been proposed so far. 
Next, we conducted experiments on artificial data to determine the conditions under which FSCPU outperforms conventional methods.
Experiments were conducted on as many as 9 baselines, including supervised, unsupervised, semi-supervised, and reinforcement learning methods.
The experiments show that FSCPU performs either competitively or superior to the conventional methods on data that satisfy the cluster assumption.
For data that satisfy the cluster assumption, FSCPU shows robust performance even when the labeling rate for positive data is reduced.
Furthermore, some conventional methods outperform FSCPU for data that do not satisfy the cluster assumption.
Finally, for the experiments on open data, we compared the performance of the downstream classification task after feature selection 
to ascertain the real-world applicability of FSCPU.
Experiments were conducted on 10 baselines and 3 variations of open data.
Experiments revealed that some variations may not satisfy the cluster assumption, but still show relatively high performance.
However, some experimental variations did not perform well, indicating that there is room for improvement in the methodology.

One limitation of FSCPU is that it performs poorly when the cluster assumption is not satisfied, as revealed in Table~\ref{tab1}.
As a future work, we will consider developing a mechanism to check in advance whether the analyzed data satisfy the cluster assumption or not.
Another case of poor performance is when the search is not sufficiently complete.
Increasing the search iteration is required, but it is not desired as FSCPU spends more wall-clock time than other approaches.\footnote{For example, for Spam dataset, FSCPU required over two hours, whereas semi-JMI finished in one minute.}
Accelerating the search strategy is an important direction of our future work.

\balance
\bibliographystyle{ACM-Reference-Format}
\bibliography{preprint}


\begin{thebibliography}{26}


\ifx \showCODEN    \undefined \def \showCODEN     #1{\unskip}     \fi
\ifx \showISBNx    \undefined \def \showISBNx     #1{\unskip}     \fi
\ifx \showISBNxiii \undefined \def \showISBNxiii  #1{\unskip}     \fi
\ifx \showISSN     \undefined \def \showISSN      #1{\unskip}     \fi
\ifx \showLCCN     \undefined \def \showLCCN      #1{\unskip}     \fi
\ifx \shownote     \undefined \def \shownote      #1{#1}          \fi
\ifx \showarticletitle \undefined \def \showarticletitle #1{#1}   \fi
\ifx \showURL      \undefined \def \showURL       {\relax}        \fi
\providecommand\bibfield[2]{#2}
\providecommand\bibinfo[2]{#2}
\providecommand\natexlab[1]{#1}
\providecommand\showeprint[2][]{arXiv:#2}

\bibitem[Baluja(1994)]%
        {pbil}
\bibfield{author}{\bibinfo{person}{Shummet Baluja}.} \bibinfo{year}{1994}\natexlab{}.
\newblock \bibinfo{booktitle}{\emph{Population-Based Incremental Learning: A Method for Integrating Genetic Search Based Function Optimization and Competitive Learning}}.
\newblock \bibinfo{type}{{T}echnical {R}eport}. \bibinfo{address}{USA}.
\newblock


\bibitem[Fan et~al\mbox{.}(2020)]%
        {Multiagent-agent_RL_2_b15}
\bibfield{author}{\bibinfo{person}{Wei Fan}, \bibinfo{person}{Kunpeng Liu}, \bibinfo{person}{Hao Liu}, \bibinfo{person}{Pengyang Wang}, \bibinfo{person}{Yong Ge}, {and} \bibinfo{person}{Yanjie Fu}.} \bibinfo{year}{2020}\natexlab{}.
\newblock \showarticletitle{AutoFS: Automated Feature Selection via Diversity-Aware Interactive Reinforcement Learning}. \bibinfo{pages}{1008--1013}.
\newblock
\href{https://doi.org/10.1109/ICDM50108.2020.00117}{doi:\nolinkurl{10.1109/ICDM50108.2020.00117}}


\bibitem[Gaudel and Sebag(2010)]%
        {single-agent_RL_2_b16}
\bibfield{author}{\bibinfo{person}{Romaric Gaudel} {and} \bibinfo{person}{Michele Sebag}.} \bibinfo{year}{2010}\natexlab{}.
\newblock \showarticletitle{Feature selection as a one-player game}. In \bibinfo{booktitle}{\emph{Proceedings of the 27th International Conference on International Conference on Machine Learning}} (Haifa, Israel) \emph{(\bibinfo{series}{ICML'10})}. \bibinfo{publisher}{Omnipress}, \bibinfo{address}{Madison, WI, USA}, \bibinfo{pages}{359–366}.
\newblock
\showISBNx{9781605589077}


\bibitem[Gheyas and Smith(2010)]%
        {Gheyas2010FeatureSS}
\bibfield{author}{\bibinfo{person}{Iffat~A. Gheyas} {and} \bibinfo{person}{Leslie~S. Smith}.} \bibinfo{year}{2010}\natexlab{}.
\newblock \showarticletitle{Feature subset selection in large dimensionality domains}.
\newblock \bibinfo{journal}{\emph{Pattern Recognition}} \bibinfo{volume}{43}, \bibinfo{number}{1} (\bibinfo{year}{2010}), \bibinfo{pages}{5--13}.
\newblock
\showISSN{0031-3203}
\href{https://doi.org/10.1016/j.patcog.2009.06.009}{doi:\nolinkurl{10.1016/j.patcog.2009.06.009}}


\bibitem[Granitto et~al\mbox{.}(2006)]%
        {DT-RFE_b7}
\bibfield{author}{\bibinfo{person}{Pablo~M. Granitto}, \bibinfo{person}{Cesare Furlanello}, \bibinfo{person}{Franco Biasioli}, {and} \bibinfo{person}{Flavia Gasperi}.} \bibinfo{year}{2006}\natexlab{}.
\newblock \showarticletitle{Recursive feature elimination with random forest for PTR-MS analysis of agroindustrial products}.
\newblock \bibinfo{journal}{\emph{Chemometrics and Intelligent Laboratory Systems}} \bibinfo{volume}{83}, \bibinfo{number}{2} (\bibinfo{year}{2006}), \bibinfo{pages}{83--90}.
\newblock
\showISSN{0169-7439}
\href{https://doi.org/10.1016/j.chemolab.2006.01.007}{doi:\nolinkurl{10.1016/j.chemolab.2006.01.007}}


\bibitem[Harik et~al\mbox{.}(1999)]%
        {cga}
\bibfield{author}{\bibinfo{person}{G.~R. Harik}, \bibinfo{person}{F.~G. Lobo}, {and} \bibinfo{person}{D.~E. Goldberg}.} \bibinfo{year}{1999}\natexlab{}.
\newblock \showarticletitle{The compact genetic algorithm}.
\newblock \bibinfo{journal}{\emph{IEEE Transactions on Evolutionary Computation}} \bibinfo{volume}{3}, \bibinfo{number}{4} (\bibinfo{date}{Nov.} \bibinfo{year}{1999}), \bibinfo{pages}{287–297}.
\newblock
\showISSN{1089-778X}
\href{https://doi.org/10.1109/4235.797971}{doi:\nolinkurl{10.1109/4235.797971}}


\bibitem[Hopkins et~al\mbox{.}(1999)]%
        {spambase_94}
\bibfield{author}{\bibinfo{person}{Mark Hopkins}, \bibinfo{person}{Erik Reeber}, \bibinfo{person}{George Forman}, {and} \bibinfo{person}{Jaap Suermondt}.} \bibinfo{year}{1999}\natexlab{}.
\newblock \bibinfo{title}{{Spambase}}.
\newblock \bibinfo{howpublished}{UCI Machine Learning Repository}.
\newblock
\newblock
\shownote{{DOI}: https://doi.org/10.24432/C53G6X}.


\bibitem[Jaskie and Spanias(2019)]%
        {PUsurvey}
\bibfield{author}{\bibinfo{person}{Kristen Jaskie} {and} \bibinfo{person}{Andreas Spanias}.} \bibinfo{year}{2019}\natexlab{}.
\newblock \showarticletitle{Positive And Unlabeled Learning Algorithms And Applications: A Survey}. In \bibinfo{booktitle}{\emph{2019 10th International Conference on Information, Intelligence, Systems and Applications (IISA)}}. \bibinfo{pages}{1--8}.
\newblock
\href{https://doi.org/10.1109/IISA.2019.8900698}{doi:\nolinkurl{10.1109/IISA.2019.8900698}}


\bibitem[Jiang et~al\mbox{.}(2019)]%
        {JSFS_b3}
\bibfield{author}{\bibinfo{person}{Bingbing Jiang}, \bibinfo{person}{Xingyu Wu}, \bibinfo{person}{Kui Yu}, {and} \bibinfo{person}{Huanhuan Chen}.} \bibinfo{year}{2019}\natexlab{}.
\newblock \showarticletitle{Joint Semi-Supervised Feature Selection and Classification through Bayesian Approach}.
\newblock \bibinfo{journal}{\emph{Proceedings of the AAAI Conference on Artificial Intelligence}} \bibinfo{volume}{33}, \bibinfo{number}{01} (\bibinfo{date}{Jul.} \bibinfo{year}{2019}), \bibinfo{pages}{3983--3990}.
\newblock
\href{https://doi.org/10.1609/aaai.v33i01.33013983}{doi:\nolinkurl{10.1609/aaai.v33i01.33013983}}


\bibitem[Ke et~al\mbox{.}(2017)]%
        {LightGBM_b22}
\bibfield{author}{\bibinfo{person}{Guolin Ke}, \bibinfo{person}{Qi Meng}, \bibinfo{person}{Thomas Finley}, \bibinfo{person}{Taifeng Wang}, \bibinfo{person}{Wei Chen}, \bibinfo{person}{Weidong Ma}, \bibinfo{person}{Qiwei Ye}, {and} \bibinfo{person}{Tie-Yan Liu}.} \bibinfo{year}{2017}\natexlab{}.
\newblock \showarticletitle{LightGBM: a highly efficient gradient boosting decision tree}. In \bibinfo{booktitle}{\emph{Neural Information Processing Systems}} (Long Beach, California, USA) \emph{(\bibinfo{series}{NIPS'17})}. \bibinfo{publisher}{Curran Associates Inc.}, \bibinfo{address}{Red Hook, NY, USA}, \bibinfo{pages}{3149–3157}.
\newblock
\showISBNx{9781510860964}


\bibitem[Khozaei and Eftekhari(2021)]%
        {MRMR-UFS_b10}
\bibfield{author}{\bibinfo{person}{Bahareh Khozaei} {and} \bibinfo{person}{Mahdi Eftekhari}.} \bibinfo{year}{2021}\natexlab{}.
\newblock \showarticletitle{Unsupervised Feature Selection Based on Spectral Clustering with Maximum Relevancy and Minimum Redundancy Approach}.
\newblock \bibinfo{journal}{\emph{International Journal of Pattern Recognition and Artificial Intelligence}} \bibinfo{volume}{35}, \bibinfo{number}{11} (\bibinfo{year}{2021}), \bibinfo{pages}{2150031}.
\newblock
\href{https://doi.org/10.1142/S0218001421500312}{doi:\nolinkurl{10.1142/S0218001421500312}}
\showeprint{https://doi.org/10.1142/S0218001421500312}


\bibitem[Liu et~al\mbox{.}(2019)]%
        {Multiagent-agent_RL_1_b14}
\bibfield{author}{\bibinfo{person}{Kunpeng Liu}, \bibinfo{person}{Yanjie Fu}, \bibinfo{person}{Pengfei Wang}, \bibinfo{person}{Le Wu}, \bibinfo{person}{Rui Bo}, {and} \bibinfo{person}{Xiaolin Li}.} \bibinfo{year}{2019}\natexlab{}.
\newblock \showarticletitle{Automating Feature Subspace Exploration via Multi-Agent Reinforcement Learning}. In \bibinfo{booktitle}{\emph{Proceedings of the 25th ACM SIGKDD International Conference on Knowledge Discovery \& Data Mining}} (Anchorage, AK, USA) \emph{(\bibinfo{series}{KDD '19})}. \bibinfo{publisher}{Association for Computing Machinery}, \bibinfo{address}{New York, NY, USA}, \bibinfo{pages}{207–215}.
\newblock
\showISBNx{9781450362016}
\href{https://doi.org/10.1145/3292500.3330868}{doi:\nolinkurl{10.1145/3292500.3330868}}


\bibitem[Liu et~al\mbox{.}(2021)]%
        {ES-MCRFS_single_agent_b9}
\bibfield{author}{\bibinfo{person}{Kunpeng Liu}, \bibinfo{person}{Pengfei Wang}, \bibinfo{person}{Dongjie Wang}, \bibinfo{person}{Wan Du}, \bibinfo{person}{Dapeng~Oliver Wu}, {and} \bibinfo{person}{Yanjie Fu}.} \bibinfo{year}{2021}\natexlab{}.
\newblock \showarticletitle{Efficient Reinforced Feature Selection via Early Stopping Traverse Strategy}. In \bibinfo{booktitle}{\emph{2021 IEEE International Conference on Data Mining (ICDM)}}. \bibinfo{pages}{399--408}.
\newblock
\href{https://doi.org/10.1109/ICDM51629.2021.00051}{doi:\nolinkurl{10.1109/ICDM51629.2021.00051}}


\bibitem[Ollivier et~al\mbox{.}(2017)]%
        {igo}
\bibfield{author}{\bibinfo{person}{Yann Ollivier}, \bibinfo{person}{Ludovic Arnold}, \bibinfo{person}{Anne Auger}, {and} \bibinfo{person}{Nikolaus Hansen}.} \bibinfo{year}{2017}\natexlab{}.
\newblock \showarticletitle{Information-Geometric Optimization Algorithms: A Unifying Picture via Invariance Principles}.
\newblock \bibinfo{journal}{\emph{Journal of Machine Learning Research}} \bibinfo{volume}{18}, \bibinfo{number}{18} (\bibinfo{year}{2017}), \bibinfo{pages}{1--65}.
\newblock
\urldef\tempurl%
\url{http://jmlr.org/papers/v18/14-467.html}
\showURL{%
\tempurl}


\bibitem[Peng et~al\mbox{.}(2005)]%
        {mRMR_b8}
\bibfield{author}{\bibinfo{person}{Hanchuan Peng}, \bibinfo{person}{Fuhui Long}, {and} \bibinfo{person}{C. Ding}.} \bibinfo{year}{2005}\natexlab{}.
\newblock \showarticletitle{Feature selection based on mutual information criteria of max-dependency, max-relevance, and min-redundancy}.
\newblock \bibinfo{journal}{\emph{IEEE Transactions on Pattern Analysis and Machine Intelligence}} \bibinfo{volume}{27}, \bibinfo{number}{8} (\bibinfo{year}{2005}), \bibinfo{pages}{1226--1238}.
\newblock
\href{https://doi.org/10.1109/TPAMI.2005.159}{doi:\nolinkurl{10.1109/TPAMI.2005.159}}


\bibitem[Rasoul et~al\mbox{.}(2021)]%
        {single-agent_RL_1_b13}
\bibfield{author}{\bibinfo{person}{Sali Rasoul}, \bibinfo{person}{Sodiq Adewole}, {and} \bibinfo{person}{Alphonse Akakpo}.} \bibinfo{year}{2021}\natexlab{}.
\newblock \bibinfo{title}{Feature Selection Using Reinforcement Learning}.
\newblock
\showeprint[arxiv]{2101.09460}~[cs.LG]
\urldef\tempurl%
\url{https://arxiv.org/abs/2101.09460}
\showURL{%
\tempurl}


\bibitem[S. and Nagapadma(2023)]%
        {rt-iot2022__942}
\bibfield{author}{\bibinfo{person}{B. S.} {and} \bibinfo{person}{Rohini Nagapadma}.} \bibinfo{year}{2023}\natexlab{}.
\newblock \bibinfo{title}{{RT-IoT2022 }}.
\newblock \bibinfo{howpublished}{UCI Machine Learning Repository}.
\newblock
\newblock
\shownote{{DOI}: https://doi.org/10.24432/C5P338}.


\bibitem[Sechidis and Brown(2018)]%
        {semi_JMI_b21}
\bibfield{author}{\bibinfo{person}{Konstantinos Sechidis} {and} \bibinfo{person}{Gavin Brown}.} \bibinfo{year}{2018}\natexlab{}.
\newblock \showarticletitle{Simple strategies for semi-supervised feature selection}.
\newblock \bibinfo{journal}{\emph{Machine Learning}} \bibinfo{volume}{107}, \bibinfo{number}{2} (\bibinfo{year}{2018}), \bibinfo{pages}{357--395}.
\newblock
\showISSN{1573-0565}
\href{https://doi.org/10.1007/s10994-017-5648-2}{doi:\nolinkurl{10.1007/s10994-017-5648-2}}


\bibitem[Sheikhpour et~al\mbox{.}(2017)]%
        {semi_survey_b2}
\bibfield{author}{\bibinfo{person}{Razieh Sheikhpour}, \bibinfo{person}{Mehdi~Agha Sarram}, \bibinfo{person}{Sajjad Gharaghani}, {and} \bibinfo{person}{Mohammad Ali~Zare Chahooki}.} \bibinfo{year}{2017}\natexlab{}.
\newblock \showarticletitle{A Survey on semi-supervised feature selection methods}.
\newblock \bibinfo{journal}{\emph{Pattern Recognition}}  \bibinfo{volume}{64} (\bibinfo{year}{2017}), \bibinfo{pages}{141--158}.
\newblock
\showISSN{0031-3203}
\href{https://doi.org/10.1016/j.patcog.2016.11.003}{doi:\nolinkurl{10.1016/j.patcog.2016.11.003}}


\bibitem[Sigillito et~al\mbox{.}(1989)]%
        {ionosphere_52}
\bibfield{author}{\bibinfo{person}{V. Sigillito}, \bibinfo{person}{S. Wing}, \bibinfo{person}{L. Hutton}, {and} \bibinfo{person}{K. Baker}.} \bibinfo{year}{1989}\natexlab{}.
\newblock \bibinfo{title}{{Ionosphere}}.
\newblock \bibinfo{howpublished}{UCI Machine Learning Repository}.
\newblock
\newblock
\shownote{{DOI}: https://doi.org/10.24432/C5W01B}.


\bibitem[Tibshirani(1996)]%
        {LASSO_b6}
\bibfield{author}{\bibinfo{person}{Robert Tibshirani}.} \bibinfo{year}{1996}\natexlab{}.
\newblock \showarticletitle{Regression Shrinkage and Selection via the Lasso}.
\newblock \bibinfo{journal}{\emph{Journal of the Royal Statistical Society. Series B (Methodological)}} \bibinfo{volume}{58}, \bibinfo{number}{1} (\bibinfo{year}{1996}), \bibinfo{pages}{267--288}.
\newblock
\showISSN{00359246}
\urldef\tempurl%
\url{http://www.jstor.org/stable/2346178}
\showURL{%
\tempurl}


\bibitem[Xu et~al\mbox{.}(2009)]%
        {Xu2009DiscriminativeSF}
\bibfield{author}{\bibinfo{person}{Zenglin Xu}, \bibinfo{person}{Rong Jin}, \bibinfo{person}{Michael~R. Lyu}, {and} \bibinfo{person}{Irwin King}.} \bibinfo{year}{2009}\natexlab{}.
\newblock \showarticletitle{Discriminative Semi-Supervised Feature Selection Via Manifold Regularization}.
\newblock \bibinfo{journal}{\emph{IEEE Transactions on Neural Networks}}  \bibinfo{volume}{21} (\bibinfo{year}{2009}), \bibinfo{pages}{1033--1047}.
\newblock
\urldef\tempurl%
\url{https://api.semanticscholar.org/CorpusID:1077768}
\showURL{%
\tempurl}


\bibitem[Xue et~al\mbox{.}(2013)]%
        {Xue2013ParticleSO}
\bibfield{author}{\bibinfo{person}{Bing Xue}, \bibinfo{person}{Mengjie Zhang}, {and} \bibinfo{person}{Will~N. Browne}.} \bibinfo{year}{2013}\natexlab{}.
\newblock \showarticletitle{Particle Swarm Optimization for Feature Selection in Classification: A Multi-Objective Approach}.
\newblock \bibinfo{journal}{\emph{IEEE Transactions on Cybernetics}} \bibinfo{volume}{43}, \bibinfo{number}{6} (\bibinfo{year}{2013}), \bibinfo{pages}{1656--1671}.
\newblock
\href{https://doi.org/10.1109/TSMCB.2012.2227469}{doi:\nolinkurl{10.1109/TSMCB.2012.2227469}}


\bibitem[Yang and Pedersen(1997)]%
        {K-best_b5}
\bibfield{author}{\bibinfo{person}{Yiming Yang} {and} \bibinfo{person}{Jan~O. Pedersen}.} \bibinfo{year}{1997}\natexlab{}.
\newblock \showarticletitle{A Comparative Study on Feature Selection in Text Categorization}. In \bibinfo{booktitle}{\emph{Proceedings of the Fourteenth International Conference on Machine Learning}} \emph{(\bibinfo{series}{ICML '97})}. \bibinfo{publisher}{Morgan Kaufmann Publishers Inc.}, \bibinfo{address}{San Francisco, CA, USA}, \bibinfo{pages}{412–420}.
\newblock
\showISBNx{1558604863}


\bibitem[Zhao and Liu(2007)]%
        {sSelect_b4}
\bibfield{author}{\bibinfo{person}{Zheng Zhao} {and} \bibinfo{person}{Huan Liu}.} \bibinfo{year}{2007}\natexlab{}.
\newblock \bibinfo{booktitle}{\emph{Semi-supervised Feature Selection via Spectral Analysis}}.
\newblock \bibinfo{pages}{641--646}.
\newblock
\href{https://doi.org/10.1137/1.9781611972771.75}{doi:\nolinkurl{10.1137/1.9781611972771.75}}
\showeprint{https://epubs.siam.org/doi/pdf/10.1137/1.9781611972771.75}


\bibitem[Zuo et~al\mbox{.}(2015)]%
        {manifold_assumptions_b17}
\bibfield{author}{\bibinfo{person}{Ling Zuo}, \bibinfo{person}{Luoqing Li}, {and} \bibinfo{person}{Chen Chen}.} \bibinfo{year}{2015}\natexlab{}.
\newblock \showarticletitle{The graph based semi-supervised algorithm with l1-regularizer}.
\newblock \bibinfo{journal}{\emph{Neurocomputing}}  \bibinfo{volume}{149} (\bibinfo{year}{2015}), \bibinfo{pages}{966--974}.
\newblock
\showISSN{0925-2312}
\href{https://doi.org/10.1016/j.neucom.2014.07.037}{doi:\nolinkurl{10.1016/j.neucom.2014.07.037}}


\end{thebibliography}

\clearpage
\appendix
\section{Proofs}\label{apdx:proof}

\begin{proof}[Proof of \Cref{prop:valid}]
Note that $\mathcal{C}_i$ are disjoint. 
Therefore, letting $c_i = \abs{\mathcal{C}_i}$ and $a_i = \abs{\mathcal{D}_{L}^\phi \cap \mathcal{C}_i}$, we have
\begin{align}
\frac{\abs{\mathcal{D}_{L}^\phi \cap \bigcup_{i \in \mathcal{K}_*}\mathcal{C}_i}^2}{ \abs{\bigcup_{i \in \mathcal{K}_*}\mathcal{C}_{i}}}
= \frac{(\sum_{i \in \mathcal{K}_*} a_i)^2}{\sum_{i \in \mathcal{K}_*} c_i}
= \frac{A^2}{C},
\end{align}
where $A = \sum_{i \in \mathcal{K}_*} a_i$ and $C = \sum_{i \in \mathcal{K}_*} c_i$. 

If $\ell \notin \mathcal{K}_*$, we must have 
\begin{align}
&\frac{(a_\ell + \sum_{i \in \mathcal{K}_*} a_i)^2}{c_\ell + \sum_{i \in \mathcal{K}_*} c_i}
< \frac{A^2}{C}
\intertext{and it is equivalent to}
\iff & \frac{(a_\ell + A)^2}{c_\ell + C}
- \frac{A^2}{C} < 0
\\
\iff & \frac{C a_\ell^2 + 2 C a_\ell A + C A^2 - C A^2 - c_\ell A^2}{(c_\ell + C) C}
< 0
\\
\iff & C a_\ell^2 + 2 C a_\ell A - c_\ell A^2 < 0
\\
\iff & \frac{a_\ell}{c_\ell }(a_\ell + 2 A)  < \frac{A^2}{C}.
\end{align}

If $k \in \mathcal{K}_*$, we must have 
\begin{align}
&\frac{(- a_k + \sum_{i \in \mathcal{K}_\phi} a_i)^2}{- c_k + \sum_{i \in \mathcal{K}_\phi} c_i}
\leq \frac{A^2}{C}.
\intertext{and it is equivalent to}
\iff & \frac{(-a_k + A)^2}{-c_k + C}
- \frac{A^2}{C} \leq 0
\\
\iff & \frac{C a_k^2 - 2 C a_k A + C A^2 - C A^2 + c_k A^2}{(-c_k + C) C}
\leq 0
\\
\iff & C a_k^2 - 2 C a_k A + c_k A^2 \leq 0
\\
\iff & \frac{a_k}{c_k }(- a_k + 2 A)  \geq \frac{A^2}{C}.
\end{align}

Altogether, we have
\begin{multline}
\frac{a_\ell}{c_\ell }(a_\ell + 2 A) < \frac{A^2}{C} \leq \frac{a_k}{c_k }(- a_k + 2 A)
\\
\implies
1 \leq \frac{(a_\ell + 2 A)}{(- a_k + 2 A)} < \frac{a_k/ c_k}{a_\ell / c_\ell }.
\end{multline}
This completes the proof.
\end{proof}

\begin{proof}[Proof of \Cref{prop:mcar}]
For a given $\mathcal{K}_*$, we have
\begin{align}
\frac{(\sum_{k \in \mathcal{K}}\abs{\mathcal{D}_{L}^\phi \cap \mathcal{C}_k})^2}{\abs{\mathcal{D}_{L}^\phi}\sum_{k \in \mathcal{K}} \abs{\mathcal{C}_k}}
&= \frac{(\sum_{i \in \mathcal{K}} c n_i \mathbb{I}\{i \leq K_A\})^2}{\abs{\mathcal{D}_L^\phi} \sum_{i \in \mathcal{K}} n_i}
\\
&= \frac{c^2}{\abs{\mathcal{D}_L^\phi} } \frac{ (\sum_{i \in \mathcal{K}} n_i \mathbb{I}\{i \leq K_A\})^2 }{ \sum_{i \in \mathcal{K}} n_i}.
\end{align}
The right-most side is maximized when $\mathcal{K}_* = \{1, \dots, K_A\}$. 
The reason is as follows. 
The inclusion of $i > K_A$ only increases the denominator. 
Therefore, $\mathcal{K}_*$ only includes $i \leq K_A$. 
Then, it is easy to see that for $\mathcal{K} \in 2^{\{1, \dots, K_A\}}$, 
\begin{equation}
\frac{c^2}{\abs{\mathcal{D}_L^\phi} } \frac{ (\sum_{i \in \mathcal{K}} n_i \mathbb{I}\{i \leq K_A\})^2 }{ \sum_{i \in \mathcal{K}} n_i}
= \frac{c^2}{\abs{\mathcal{D}_L^\phi} } \sum_{i \in \mathcal{K}} n_i
\end{equation}
and the right-hand side is maximized when $\mathcal{K} = \{1, \dots, K_A\}$. This completes the proof.
\end{proof}

\section{Additional Results }
\begin{figure*}
    \centering
    \begin{subfigure}[t]{0.25\textwidth} 
        \centering
        \includegraphics[width=\linewidth]{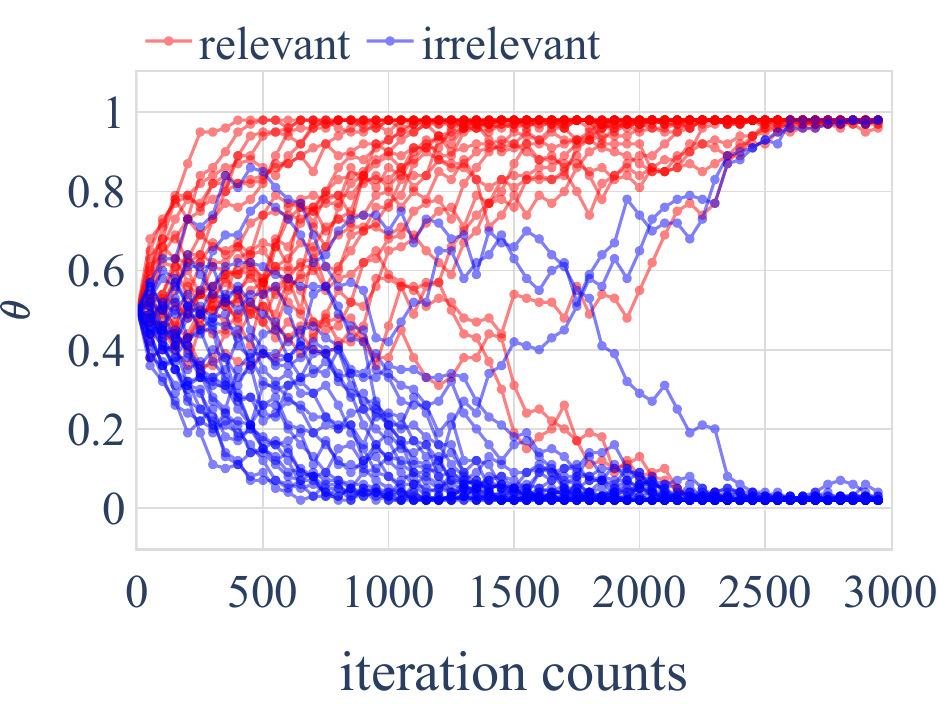}
        \caption{conditions = \{$\checkmark$, 40\%, 8, 1\}} 
    \end{subfigure}%
    \begin{subfigure}[t]{0.25\textwidth}
        \centering
        \includegraphics[width=\linewidth]{./FSCPU_d2.pdf}
        \caption{conditions = \{$\checkmark$, 40\%, 8, 2\}}
    \end{subfigure}%
    \begin{subfigure}[t]{0.25\textwidth}
        \centering
        \includegraphics[width=\linewidth]{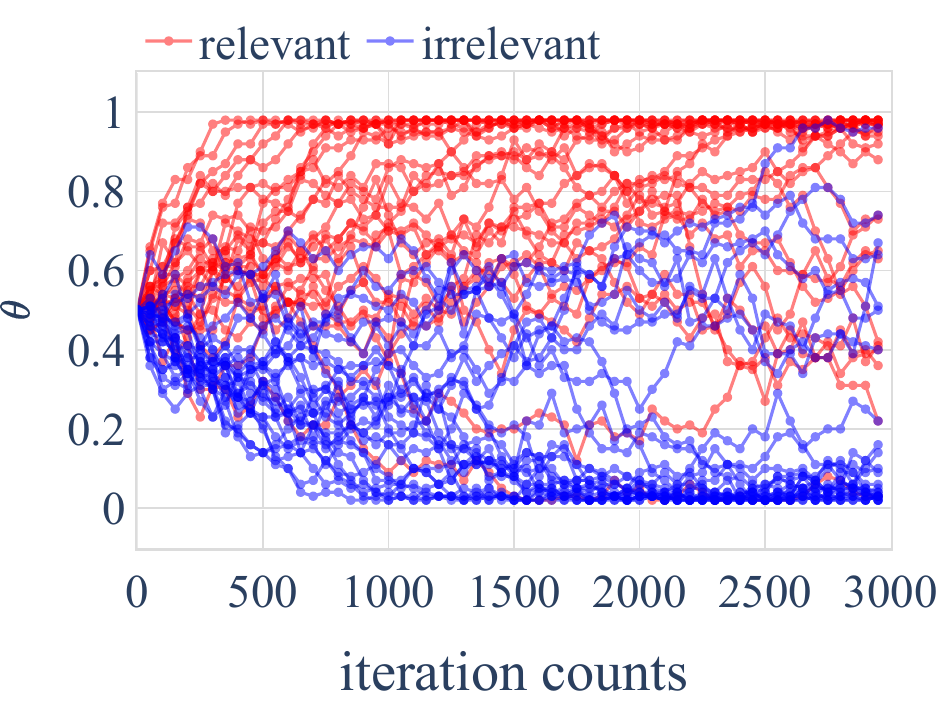}
        \caption{conditions = \{$\checkmark$, 10\%, 8, 1\}}
    \end{subfigure}%
    \begin{subfigure}[t]{0.25\textwidth}
        \centering
        \includegraphics[width=\linewidth]{./FSCPU_d4.pdf}
        \caption{conditions = \{$\checkmark$, 10\%, 8, 2\}}
    \end{subfigure}%

    \centering
    \begin{subfigure}[t]{0.25\textwidth} 
        \centering
        \includegraphics[width=\linewidth]{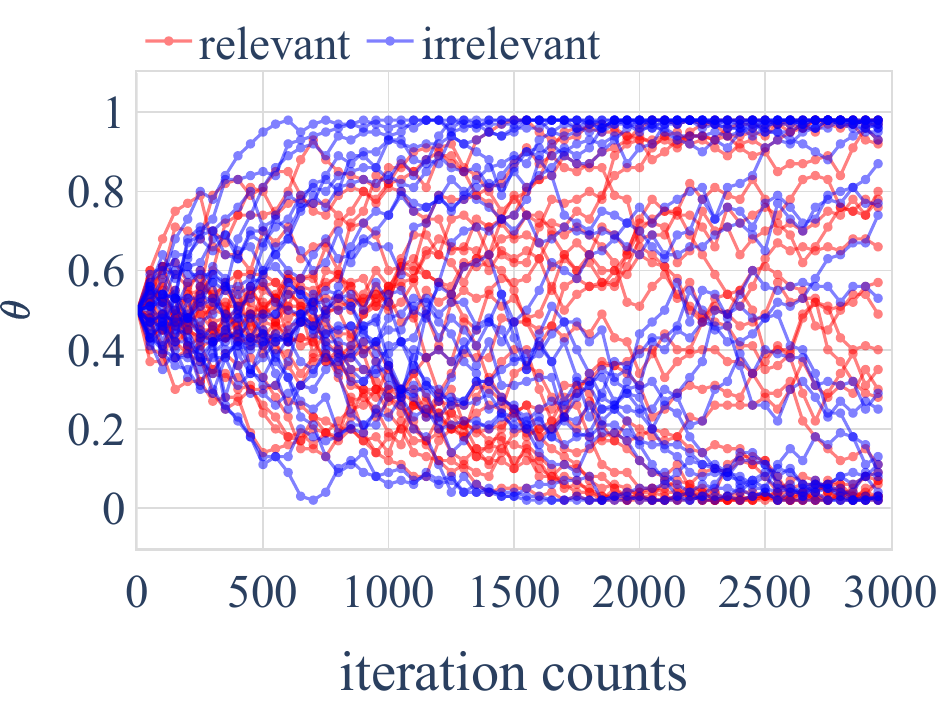}
        \caption{conditions = \{$\times$, 40\%\}} 
    \end{subfigure}%
    \begin{subfigure}[t]{0.25\textwidth}
        \centering
        \includegraphics[width=\linewidth]{./FSCPU_d6.pdf}
        \caption{conditions = \{$\times$, 10\%\}}
    \end{subfigure}%
    \begin{subfigure}[t]{0.25\textwidth}
        \centering
        \includegraphics[width=\linewidth]{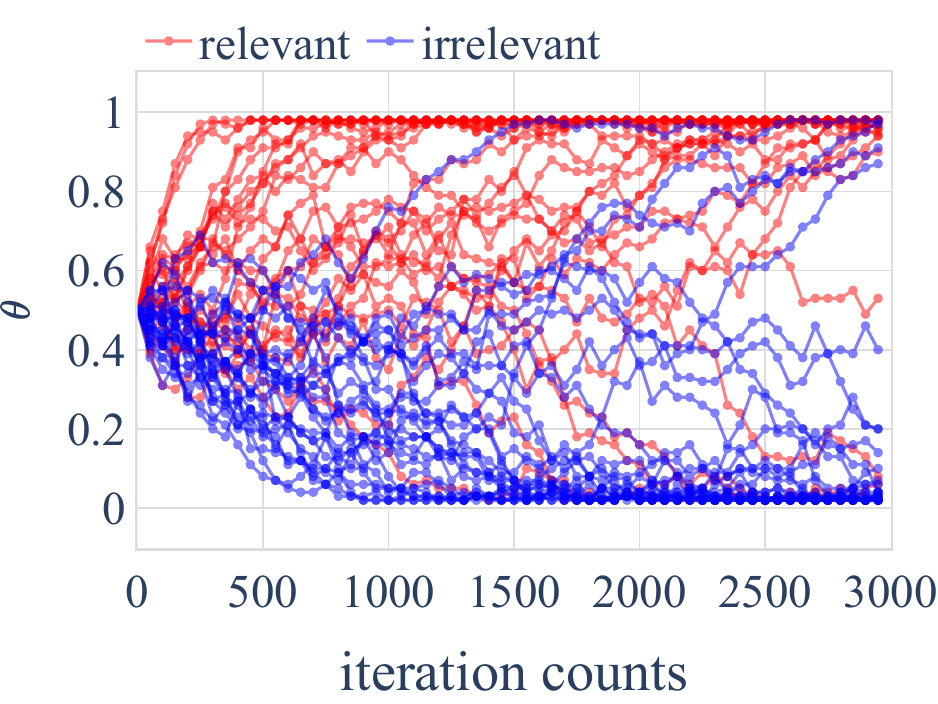}
        \caption{conditions = \{$\checkmark$, 40\%, 1, 1\}}
    \end{subfigure}%
    \begin{subfigure}[t]{0.25\textwidth}
        \centering
        \includegraphics[width=\linewidth]{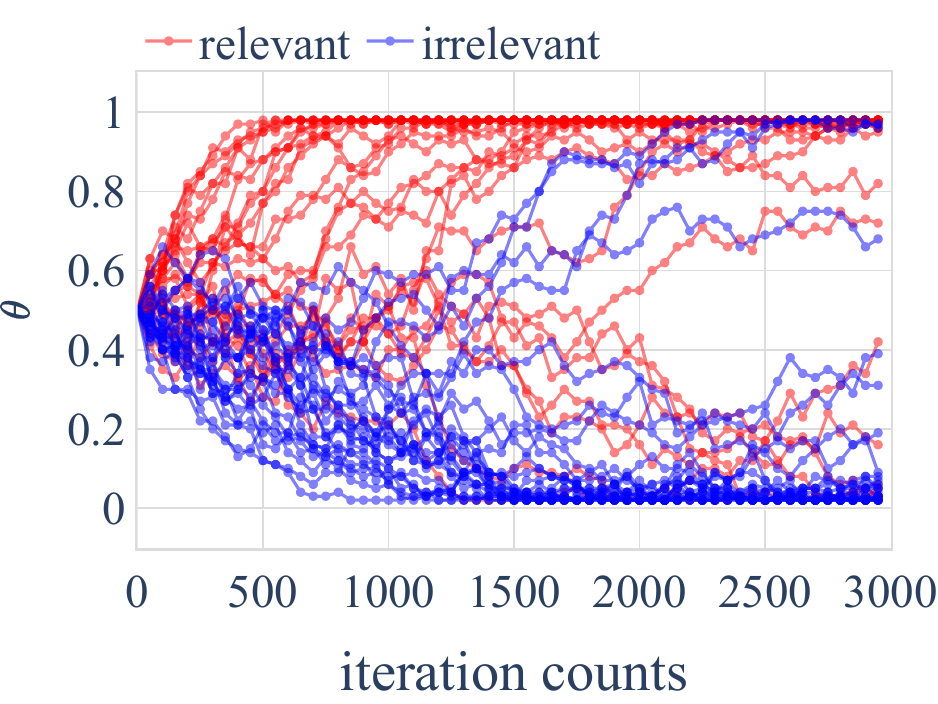}
        \caption{conditions = \{$\checkmark$, 40\%, 1, 2\}}
    \end{subfigure}%

    \centering
    \begin{subfigure}[t]{0.25\textwidth} 
        \centering
        \includegraphics[width=\linewidth]{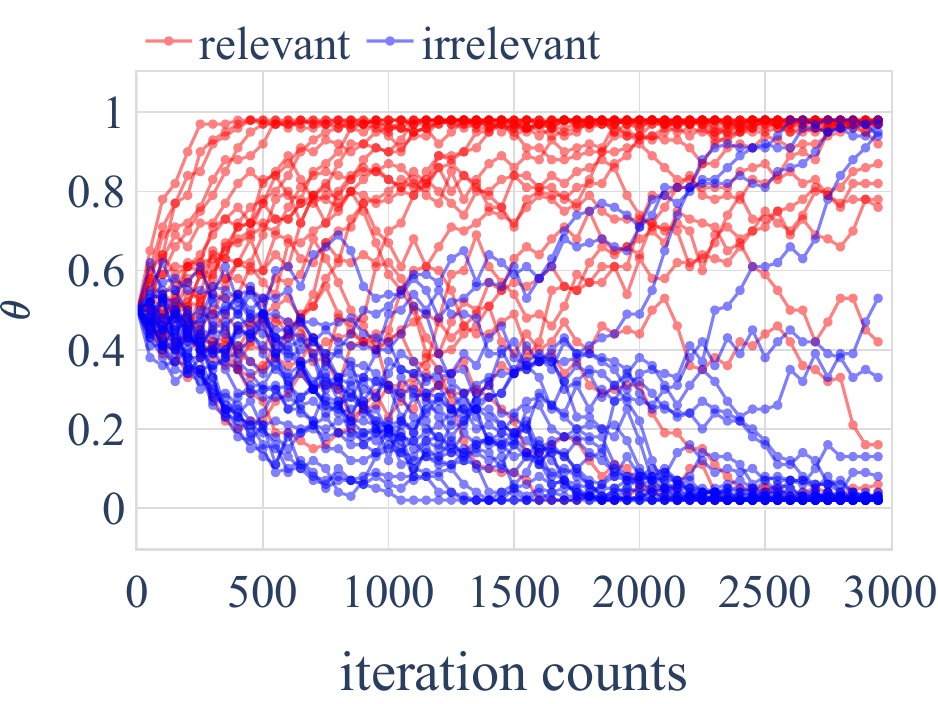}
        \caption{conditions = \{$\checkmark$, 10\%, 1, 1\}} 
    \end{subfigure}%
    \begin{subfigure}[t]{0.25\textwidth}
        \centering
        \includegraphics[width=\linewidth]{./FSCPU_d10.pdf}
        \caption{conditions = \{$\checkmark$, 10\%, 1, 2\}}
    \end{subfigure}%
    \begin{subfigure}[t]{0.25\textwidth}
        \centering
        \includegraphics[width=\linewidth]{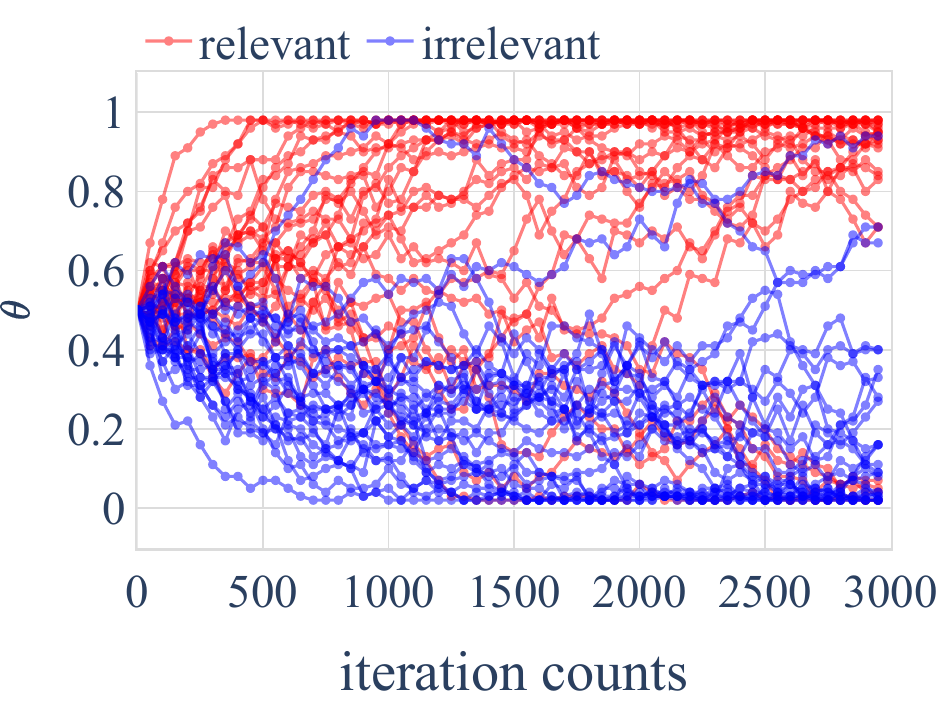}
        \caption{conditions = \{$\checkmark$, 40\%, 8, 1\}}
    \end{subfigure}%
    \begin{subfigure}[t]{0.25\textwidth}
        \centering
        \includegraphics[width=\linewidth]{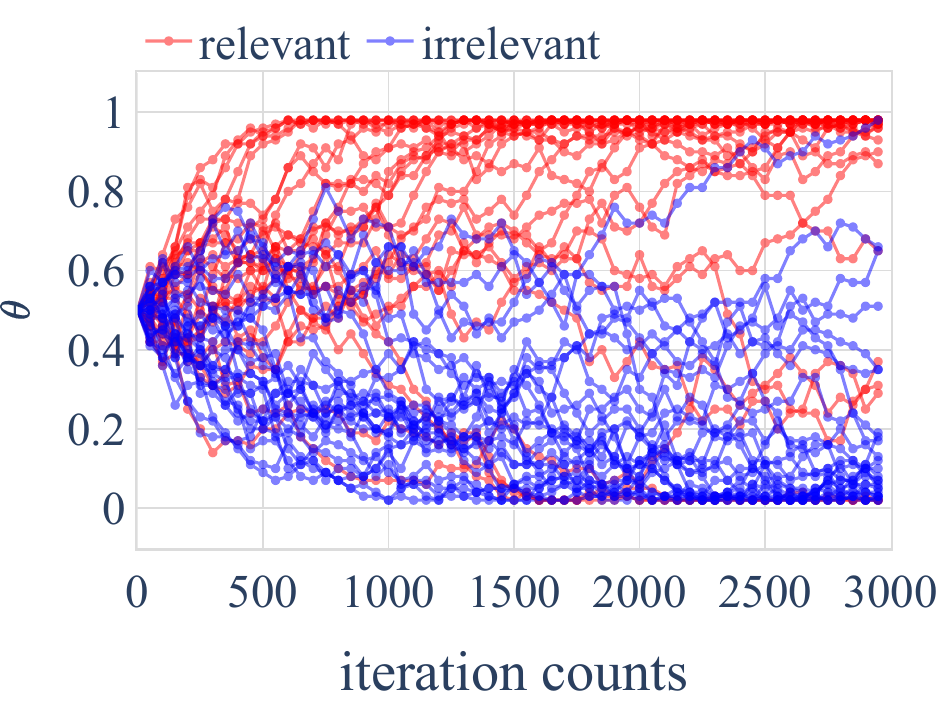}
        \caption{conditions = \{$\checkmark$, 40\%, 8, 2\}}
    \end{subfigure}%
    
    \centering
    \begin{subfigure}[t]{0.25\textwidth} 
        \centering
        \includegraphics[width=\linewidth]{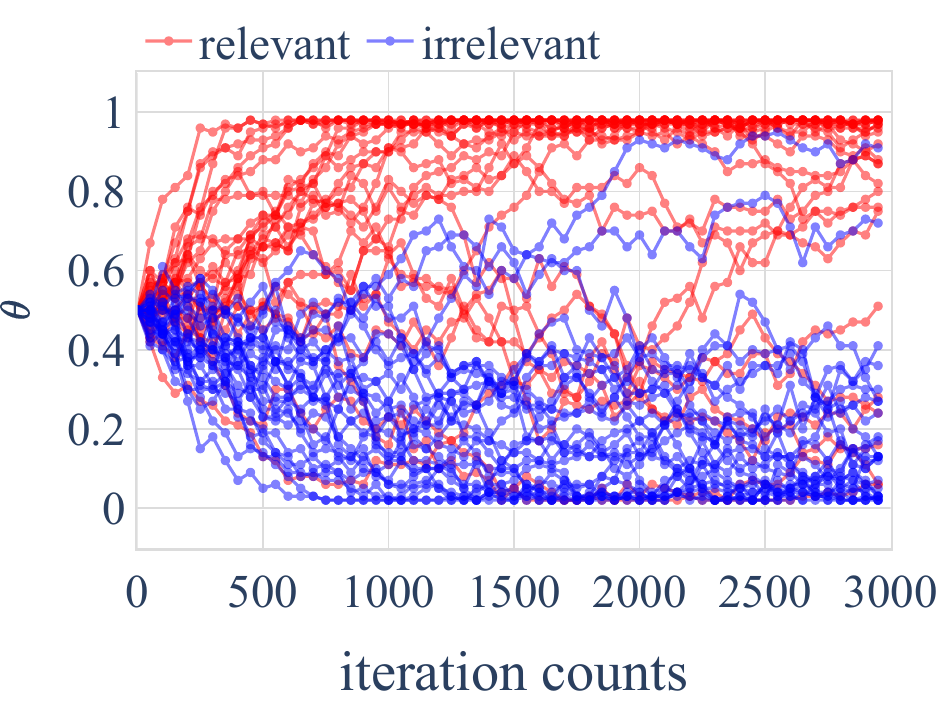}
        \caption{conditions = \{$\checkmark$, 10\%, 8, 1\}} 
    \end{subfigure}%
    \begin{subfigure}[t]{0.25\textwidth}
        \centering
        \includegraphics[width=\linewidth]{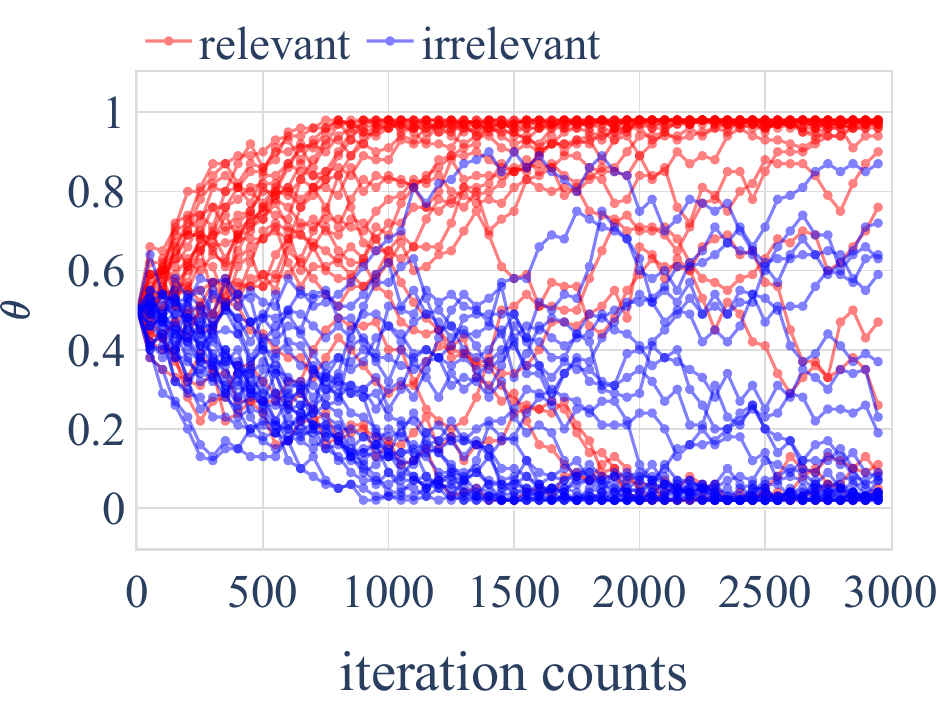}
        \caption{conditions = \{$\checkmark$, 10\%, 8, 2\}}
    \end{subfigure}%
    \begin{subfigure}[t]{0.25\textwidth}
        \centering
        \includegraphics[width=\linewidth]{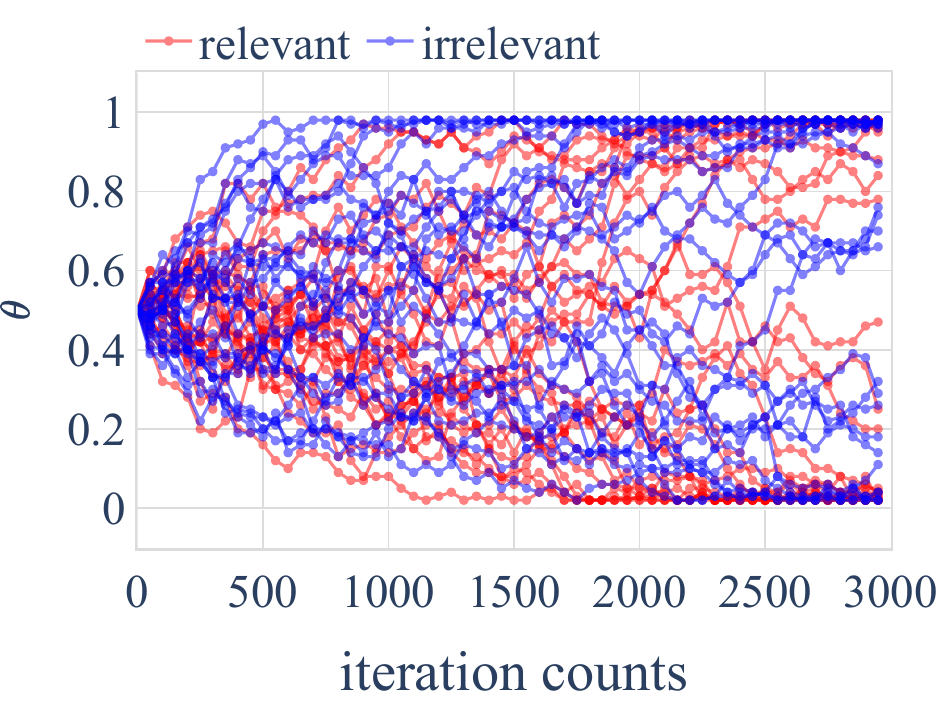}
        \caption{conditions = \{$\times$, 40\%\}}
    \end{subfigure}%
    \begin{subfigure}[t]{0.25\textwidth}
        \centering
        \includegraphics[width=\linewidth]{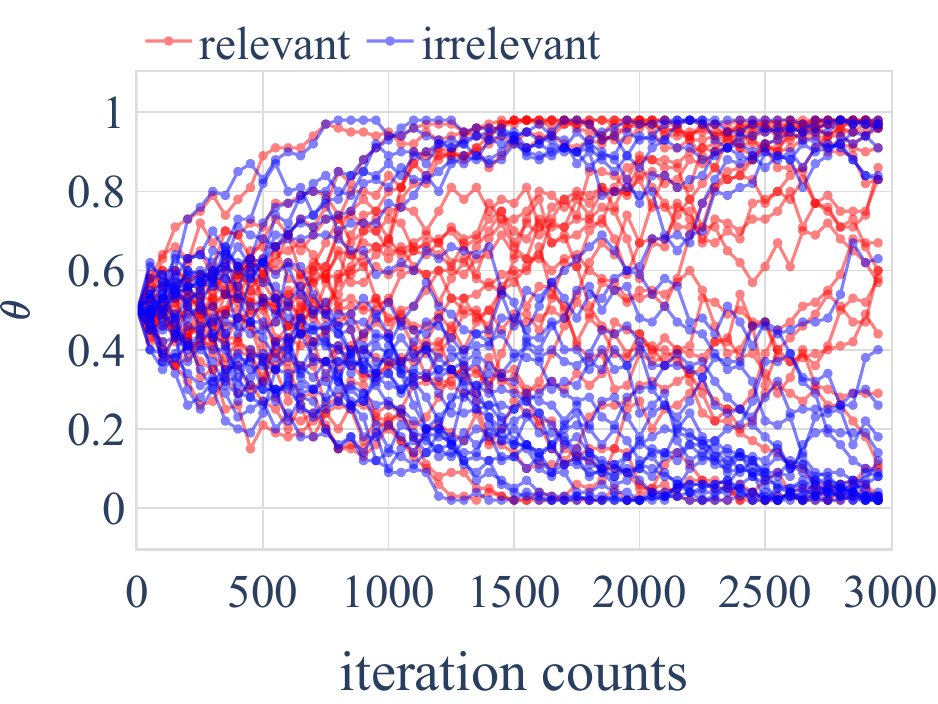}
        \caption{conditions = \{$\times$, 10\%\}}
    \end{subfigure}%
    \label{fig:Additional4}

    \centering
    \begin{subfigure}[t]{0.25\textwidth} 
        \centering
        \includegraphics[width=\linewidth]{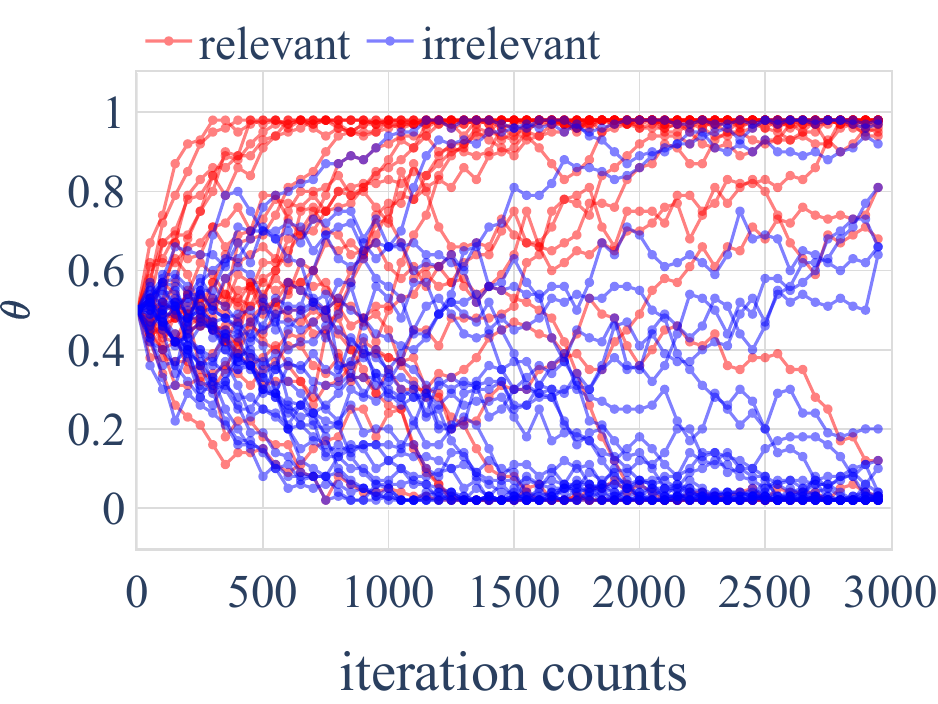}
        \caption{conditions = \{$\checkmark$, 40\%, 1, 1\}} 
    \end{subfigure}%
    \begin{subfigure}[t]{0.25\textwidth}
        \centering
        \includegraphics[width=\linewidth]{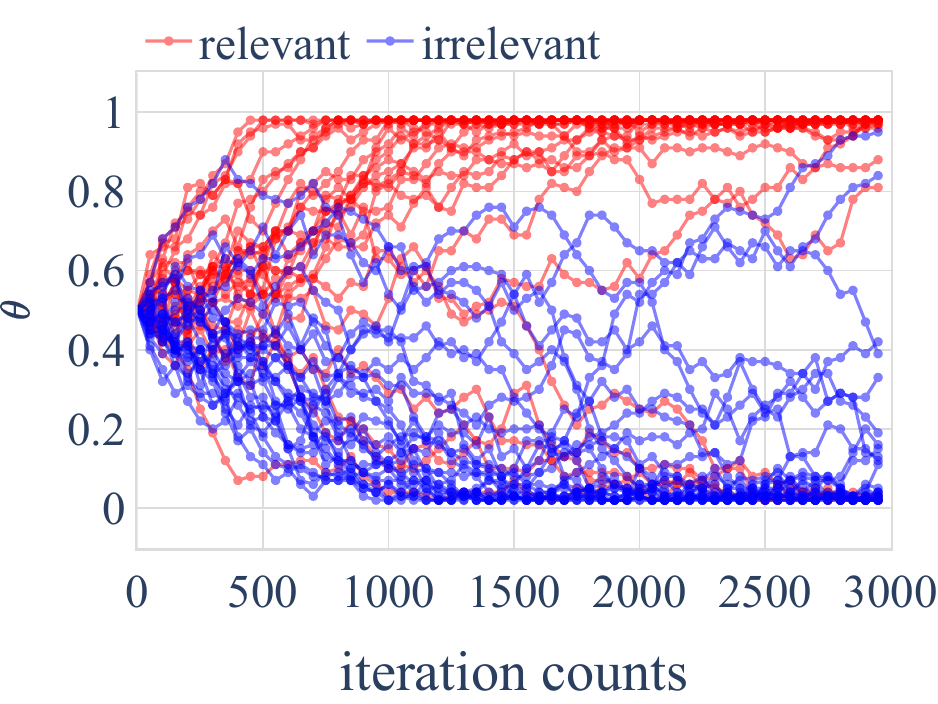}
        \caption{conditions = \{$\checkmark$, 40\%, 1, 2\}}
    \end{subfigure}%
    \begin{subfigure}[t]{0.25\textwidth}
        \centering
        \includegraphics[width=\linewidth]{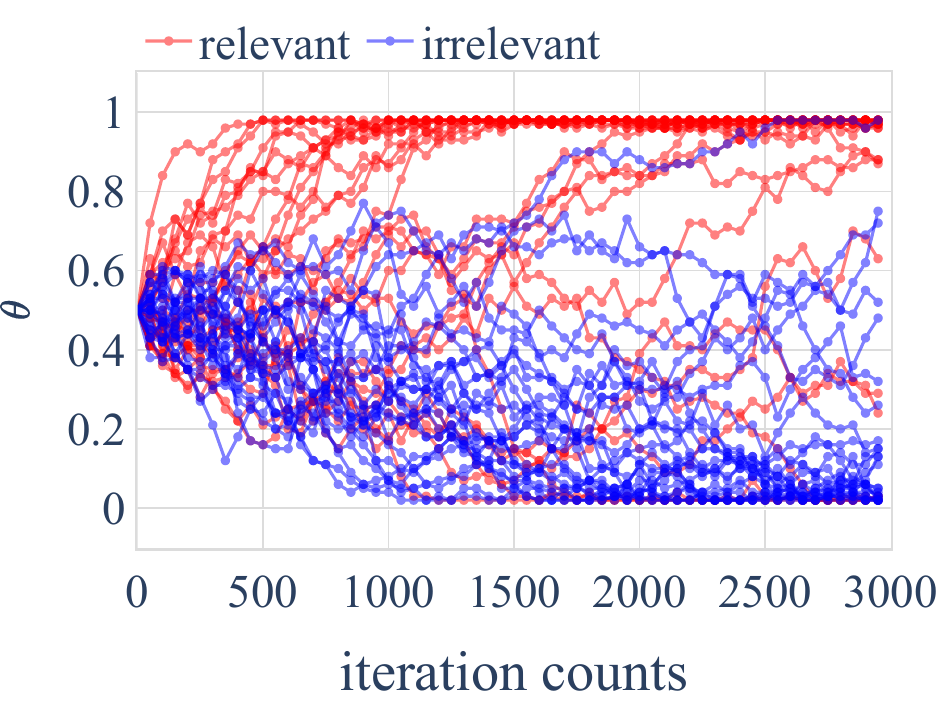}
        \caption{conditions = \{$\checkmark$, 10\%, 1, 1\}}
    \end{subfigure}%
    \begin{subfigure}[t]{0.25\textwidth}
        \centering
        \includegraphics[width=\linewidth]{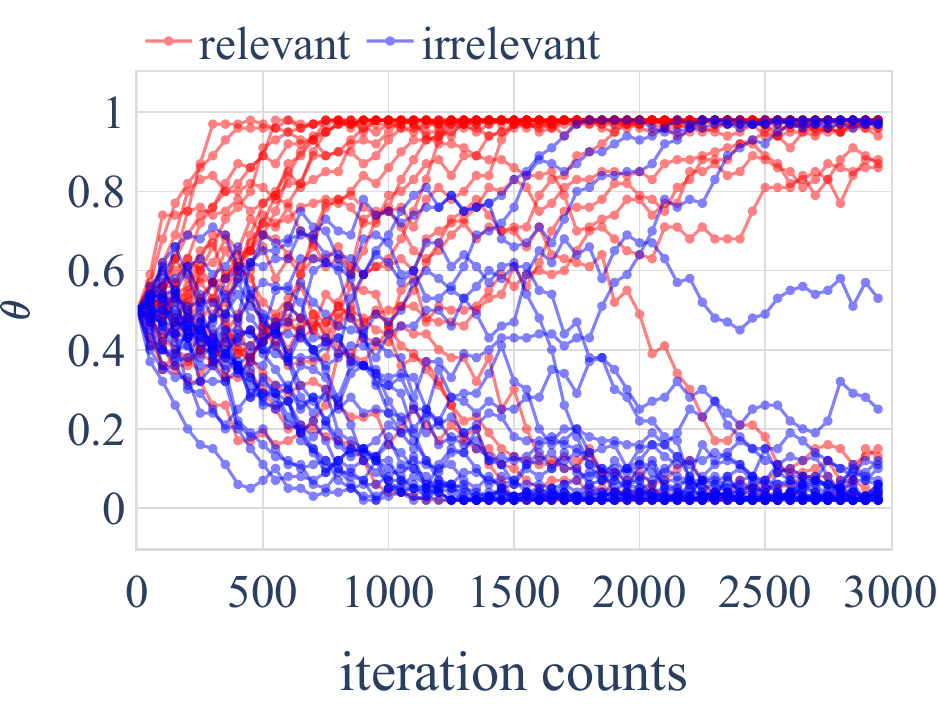}
        \caption{conditions = \{$\checkmark$, 10\%, 1, 2\}}
    \end{subfigure}%
    \caption{Convergence processes of stochastic parameters in experiments with artificial data (all variations). (a)-(j): FSCPU results, (k)-(t): FSCPU-MI results.}
    \label{fig:Additional}
\end{figure*}

Figure \ref{fig:Additional} summarizes the convergence processes for all experiments conducted on artificial data in Section \ref{ExperimentsArtificialData}.

\section{Implementation details}
Table \ref{tab:libraries} lists the library versions and methods used in the implementation of this study.
\begin{table}[h]\small
\centering
\caption{Summary of libraries and methods used in the experiments.}
\label{tab:libraries}
\begin{tabular}{lll}
\toprule
\textbf{Library}      & \textbf{Version} & \textbf{Methods and Models} \\ 
\midrule
scikit-learn          & 1.0.2            & 
\begin{tabular}[c]{@{}l@{}}
SelectKBest: K-Best \\ 
chi2: K-Best \\ 
Lasso: Lasso \\ 
KMeans: MRMR-UFS \\ 
RandomForestClassifier: DT-RFE, ES-MCRFS \\ 
NearestNeighbors: sSelect, JSFS, MRMR-UFS \\ 
normalized\_mutual\_info\_score: sSelect
\end{tabular} \\ 
\midrule
Keras                 & 2.2.4            & ES-MCRFS \\ 
\midrule
lightgbm              & 3.3.2            & LGBM-OPT \\ 
\midrule
pymrmr                & 0.1.11           & mRMR \\ 
\bottomrule
\end{tabular}
\end{table}

\end{document}